\documentclass[twoside]{article}

\usepackage[accepted]{aistats2023}
\usepackage{microtype}
\usepackage{graphicx}
\usepackage{subcaption}

\PassOptionsToPackage{hyphens}{url}\usepackage{hyperref}
\usepackage{booktabs} 
\usepackage{bookmark}

\usepackage{amsmath}
\usepackage{amssymb}
\usepackage{mathtools}
\usepackage{amsthm}
\usepackage{algorithmicx}
\usepackage{tabularx}

\usepackage[linesnumbered,ruled,vlined]{algorithm2e}
\usepackage{parskip}

\usepackage[capitalize,noabbrev]{cleveref}
\usepackage{natbib}

\theoremstyle{plain}

\theoremstyle{definition}

\theoremstyle{remark}

\usepackage[textsize=tiny]{todonotes}

\begin{document}

\runningauthor{T. Kim, S. Singh, N. Madaan, C. Joe-Wong}

\twocolumn[
\aistatstitle{Characterizing Internal Evasion Attacks in Federated Learning}
\vspace{-4mm}
\aistatsauthor{Taejin Kim \And Shubhranshu Singh}
% \aistatsaddress{Carnegie Mellon University} 
\aistatsaddress{Carnegie Mellon University \And Carnegie Mellon University} 
\aistatsauthor{Nikhil Madaan \And Carlee Joe-Wong }
\aistatsaddress{Carnegie Mellon University \And Carnegie Mellon University} 
]

\begin{abstract}
Federated learning allows for clients in a distributed system to jointly train a machine learning model. However, clients' models are vulnerable to attacks during the training and testing phases. In this paper, we address the issue of adversarial clients performing ``internal evasion attacks'':  crafting evasion attacks at test time to deceive other clients. For example, adversaries may aim to deceive spam filters and recommendation systems trained with federated learning for monetary gain. The adversarial clients have extensive information about the victim model in a federated learning setting, as weight information is shared amongst clients. We are the first to characterize the transferability of such internal evasion attacks for different learning methods and analyze the trade-off between model accuracy and robustness depending on the degree of similarities in client data. We show that adversarial training defenses in the federated learning setting only display limited improvements against internal attacks. However, combining adversarial training with personalized federated learning frameworks increases relative internal attack robustness by 60\% compared to federated adversarial training and performs well under limited system resources. 
\end{abstract}

\section{INTRODUCTION}
\label{sec:intro}

Modern computing devices such as smartphones 
% and IoT (Internet of Things) sensors 
collect increasing amounts of data to improve user applications and services. Much of this data, however, is privacy-sensitive (e.g., health data from biological sensors or smartphone usage data). Federated learning has emerged as a distributed training paradigm \citep*{FLmobile, fedlearn_survey} that allows multiple users to collectively train a model, without revealing sensitive data to each other.
The growth of federated learning, and machine learning in general, has allowed such algorithms to improve many applications, but it has also fueled attacks on learning algorithms. Evasion attacks~\citep*{biggio2013evasion, madryAttacks}, for example, aim to perturb inputs to trained models that are undetectable to human users but change the model output at test time. Slightly altering a stop sign, for example, might lead to it being classified as a speed limit sign instead~\citep*{cao2017mitigating}. Such attacks can endanger users by subverting their ability to trust the outputs of trained models. 

\begin{figure}[t!]
\begin{center}
\centerline{\includegraphics[width=0.75\columnwidth]{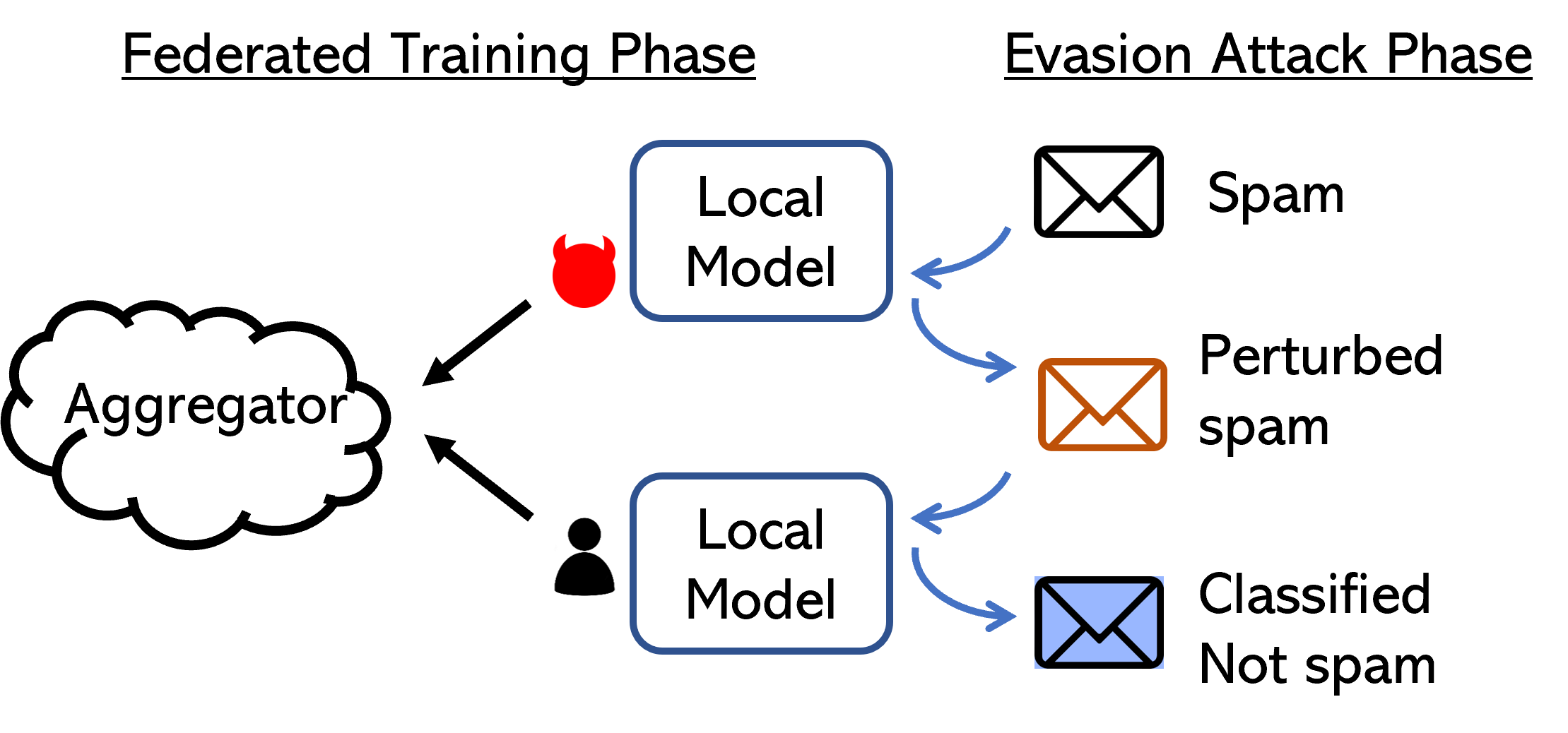}}
    \vskip -0.1in
\caption{\small{Within a federated learning setting, an internal evasion attack is performed by an adversarial participant that generates an evasion attack using the overlap between its local model and victim participant's model to bypass a spam filter.}}
\label{fig:internal_atk}
\end{center}
\vskip -0.35in
\end{figure}

\textbf{Threat Model.}
For an \emph{internal} evasion attack in a federated learning setting, we suppose that adversaries can access the models of compromised clients (e.g., by participating in the federated learning process). They can then generate adversarial perturbations on their own trained models at test time and feed the perturbed data to other clients. For example, email spam filters can be trained through federated learning, and a malicious client may use its local spam filter model to craft messages that can bypass the filters of other clients \citep*{spam_filter}, as seen in Figure \ref{fig:internal_atk}.  Such attacks 
can not be detected during training time by other clients or servers,
since the attack can be computed entirely locally post-training. Furthermore,  federated learning in particular may suffer from internal adversaries due to the many (sometimes thousands) of clients involved. 
We further analyze the threat model of \emph{external} adversaries, where the adversary generating evasion attacks does not participate in the federated learning process, and generates evasion attacks based on a substitute model trained separately.

\begin{table}[t]
\centering
\caption{\small{Amount of the adversary's information regarding victim federated clients given federated learning participation (internal or external) and training method.}}
\begin{tabular}{l|lll}
\hline
Adv.\textbackslash{}Method & Local     & FedAvg    & FedEM     \\ \hline
Internal Adv.              & Black-box & White-box & Grey-box  \\
External Adv.              & Black-box & Black-box & Black-box \\ \hline
\end{tabular}
\label{tbl:threat_box}
\vskip -0.15in
\end{table}

While external adversaries only have limited black-box information on victim federated clients, internal adversaries have different amounts of information of victim clients' models based on the method of training, as seen in Table \ref{tbl:threat_box}.
The traditional
federated learning method, where all participating clients train a single model through aggregation,
leads to a white-box situation as all clients including the internal adversary have the same model.
Locally trained models may reduce the transferability of internal attacks, as no information is shared during the training process leading to a black-box situation. 
We thus observe a \emph{trade-off between accuracy and robustness to evasion attacks}, as federated learning's collaboration between clients leads to higher test accuracy, but the adversary then has more information about the victim to generate highly effective evasion attacks.
Personalized federated learning has clients collaboratively train models,
but personalize them to individual data distributions.
Such a method, that leaves adversaries only with grey-box information on victim models, can transcend this trade-off: clients can make use of each others' information to improve their models, but retain enough differences between their models to remain robust against internal evasion attacks. Although the high level intuition of this accuracy to robustness trade-off is clear, it is unclear how this trade-off will manifest quantitatively, especially when using personalized federated learning as a foundation of defense. The use of personalized federated learning also poses a new research question:
\emph{how can we control differences between clients' models so as to maintain high accuracy, with minimal internal attack transferability?}

Existing defenses against evasion attacks in federated learning generally utilize adversarial training~\citep*{FAT, biasvar, certifiedfed, chencertifiablyRobust}, where clients generate adversarial inputs and incorporate them into the training process, which has been shown to be an effective and reliable defense method against evasion attacks \citep*{madryAttacks}. However, in the federated learning setting, the attacking client has full knowledge of the victim's model (white-box setting) when performing internal attacks, with adversarial training demonstrating poor performance against internal evasion attacks when compared to attacks by external adversaries (black-box setting). Thus, adversarial training on its own does not sufficiently reduce the success of internal evasion attacks. By utilizing personalized federated learning together with adversarial training, the adversary does not have white-box information of the victim models, leading to a more effective robustness gain through adversarial training.

In this paper, our main \textbf{contributions} are to characterize and measure internal evasion attacks in federated learning. We first demonstrate an adversarial client's ability to perform internal evasion attacks to other clients in a federated learning system. We then explore the effectiveness of adversarial training and personalized federated learning as a defense mechanism against internal attacks.
In doing so, we make the following technical contributions:

\begin{itemize}
    \item We are the first, to the best of our knowledge, to \emph{characterize and analyze the novel threat model of internal evasion attacks} between federated learning models.  
    We quantify a trade-off between accuracy and robustness to internal and external evasion attacks when comparing different types of federated learning to local learning.
    \item We experimentally show that existing defenses, namely federated adversarial training, are unable to provide sufficient robustness against internal evasion attacks.
    \item We introduce a novel defense, pFedDef, that combines personalized learning with adversarial training. Our \emph{extensive experiments} show that pFedDef, whilst respecting heterogeneous client resource constraints, maintains high test accuracy and robustness to external attacks while increasing robustness against internal evasion attacks compared to existing methods on the CIFAR-10, CelebA, and the GNN Fake News data sets, 
    with at least a relative robustness gain of 60\%.
\end{itemize}

The remainder of this paper is organized as follows. Section \ref{sec:related_works} contrasts our analysis of internal evasion attacks with related works. 
Section \ref{sec:transferability} characterizes such attacks within different federated learning schemes corresponding to white-, black-, and grey-box models. Section \ref{sec:FAT} evaluates the internal attack robustness of existing federated adversarial training methods. We introduce pFedDef in Section \ref{sec:pFedDef}. 
Different methods of personalization and adversarial threats are further examined in Section \ref{sec:exp}. We conclude in Section \ref{sec:conc}.

\textbf{Assumptions and Societal Impact.} We utilize FedEM \citep*{fedMTL21} to depict personalized federated learning, which has specific assumptions on the underlying data distributions amongst clients and is resource intensive. The pFedDef algorithm as currently presented utilizes a measurement of data distribution similarity amongst clients that is unique to FedEM. However,  
different personalization methods can be used (e.g., local tuning) with appropriate measurements of client data similarity to remove the assumption on the data distribution and to reduce overhead. All evasion attacks analyzed in this study are limited to untargeted and gradient-based attacks, although we expect our results to generalize to other attacks due to model inter-boundary distance measurements, which implies similar trends across different attack types.
We note that the attack model in this work may have a negative societal impact, as we are the first to analyze the internal attack scenario in federated learning.

%%%% Condensed Version Below
\section{RELATED WORKS}
\label{sec:related_works}

\citet*{FAT} first introduced evasion attacks against a model that is trained through federated learning. Here, adversarial training is performed alongside federated learning to increase robustness against evasion attacks. However, we show that the robustness gained through adversarial training is less effective
against \emph{internal} evasion attacks when compared to \emph{external} attacks. Furthermore, limited hardware and communication resources, as well as the non-i.i.d. (independent and identically distributed) nature of clients' data sets, impedes both model accuracy and performance during adversarial training \citep*{commconstrain,affineshift}.  Further extensions have proposed methods of incorporating the impact of non-i.i.d. data on federated adversarial training, e.g., sending centrally generated adversarial examples to clients based on the federated learning generalization error~\citep*{biasvar}, or using batch normalization to propagate adversarial robustness from nodes with resources to nodes without~\citep*{bnprop}.
Other approaches have attempted to improve federated adversarial robustness, including randomized smoothing of models \citep*{chencertifiablyRobust}. Our work in comparison analyzes and addresses the issue of heterogeneous data and hardware capabilities through the use of \emph{personalized} federated learning, which raises new questions on how adversarial training samples may propagate between clients with different models. 
Our work examines model defenses against multi-step attacks from other clients who have grey-box information of the victim models, which are more potent than black-box or single-step white-box attacks \citep*{REN2020346}.

Another line of work focuses on \emph{Sybil attacks} and \emph{robust aggregation schemes} of federated user updates. Krum, Bulyan, and trimmed-mean aggregation methods \citep*{Krum, diffprivacy, trimmedmean, foolsgold}, for instance, are proposed defenses against data and model poisoning attacks (e.g., label flipping).
Works regarding Sybil attacks solely focus on the training phase of federated learning, while our defense mechanism pFedDef defends against evasion attacks occurring at the test phase. Training-based defenses designed to detect and neutralize poisoned models or data may not be effective against such internal evasion attacks.
However, we observe in Section \ref{sec:exp} that pFedDef does increase robustness to label flipping attacks that occur in the training phase. The work in \citet*{FAT} examines the impact of robust aggregation schemes on traditional federated adversarial training, while \citet*{certifiedfed} introduce a ``backdoor'' poisoning attack that spans both the training and testing phases. These are distinct from internal evasion attacks.

Outside the federated learning context, \emph{ensemble attacks} \citep*{bagboostensemble} generate evasion attacks by combining gradient information from different models in order to increase transferability of attacks through generalization. We evaluate pFedDef on ensemble internal attacks in Section~\ref{sec:exp}. Ensemble attacks leverage the diversity present in ensembles of models, similar to the different client models present in the personalized federated learning setting.

This work has been presented as an abstract at the CrossFL Workshop 2022 \citep*{ref:crossFL_tj}. The pFedDef library used for experimentation in this paper has been published in the Software Impacts journal \citep*{software_impacts}.

\section{CHARACTERIZING EVASION ATTACK TRANSFERABILITY IN FEDERATED LEARNING}
\label{sec:transferability}

During the aggregation phase of federated learning, information is shared between clients, increasing the trained model's accuracy due to training over an expanded and more diverse data set \citep*{FedAvg}. However, as models of clients become more similar (or identical) due to the aggregation process, adversarial clients have more information on the victim's model, allowing the crafting of more effective internal evasion attacks.
This section defines the internal evasion attack, contrasting it to the external evasion attack, and empirically characterizes the trade-off of accuracy to robustness by comparing federated learning to local learning. Some methods, such as personalized federated learning or local tuning, bridge the gap between local and federated learning by allowing clients to share information amongst themselves while maintaining differences between their models. These methods are further examined in Section~\ref{sec:pFedDef} and Section~\ref{sec:exp}, respectively.

\subsection{Crafting Internal Evasion Attacks}

The aim of \emph{evasion attacks} is to alter the input $x$ to alter the model prediction for $x$.  
The projected gradient descent (PGD) method is one of the most popular and effective forms of evasion attacks \citep*{madryAttacks, REN2020346}. 
The PGD method is used for the illustration of evasion attacks and adversarial training in this paper.
In such attacks, the adversary, 
aiming to induce any incorrect classification label, iteratively updates the current input $x^t$ as:
\begin{align}\label{PGD}
    x^{t+1} = \Pi_{x+S} \left(x^t + \alpha_{\text{PGD}} \text{sgn}(\nabla_x L(h_{\theta}, x,y))\right)
\end{align}
The input $x$ with correct label $y$ is perturbed along the gradient of the model loss function $L$ with weights $h_{\theta}$. The step size $\alpha_{\text{PGD}}$ is chosen to not be too small so that an effective perturbation can be quickly found, while not too large such that effective perturbations are not omitted. The perturbation to input $x$ is then projected $(\Pi_{x + S})$ to be within the perturbation budget $S$. The perturbation budget exists such that perturbations are not obvious to detection (e.g., a heavily perturbed image may be noticed by the human eye). This budget is most often a $l_2$ or $l_\infty$ norm-ball.

For external evasion attacks, as considered in prior works~\citep*{advtransfer, advFAIL}, the adversary must train a substitute model $g_\theta$ to imitate the victim model $h_\theta$, and craft adversarial examples following Equation~\ref{PGD} on the substitute model. In comparison, our work examines \emph{internal} evasion attacks, where clients participating in the federated learning process craft evasion attacks during test time. In traditional federated learning, such clients need no substitute model, as the victim model weight parameters $h_{\theta}$ are known by the adversary,
which makes evasion attacks more potent as adversarial perturbations are generated with respect to the gradient information of the victim model. Local training on each client's data, in contrast, leads to client-specific models $h_{\theta_c}$; thus, even internal adversarial clients would not have access to the victim client's model, leading to a black-box attack situation, where models $h_{\theta_c}$ acts as a substitute model for attack generation.

\subsection{Evaluation of Internal Evasion Attacks on Federated Learning} \label{sec:trans_eval}

We empirically compare the threats of external and internal evasion attacks by evaluating their success on FedAvg (all clients train the same model), local training (each client trains a model on its data alone), and FedEM~\citep*{fedMTL21}, a personalized method (see Section~\ref{sec:pFedDef}).
We run each of these three learning methods on the CIFAR-10, Celeba \citep*{leaf} and Fake News \citep*{fake_news_set, dou2021user} data sets. 
The data are split in a non-i.i.d. fashion amongst 40 clients. The MobileNetV2 architecture is used for the first two data sets, while a multi-layer perceptron is used for the Fake News data set, which aims to perform a binary classification task on text representation of news articles. The training rates are $\alpha = 0.03$ for local and personalized federated learning and $\alpha = 0.01$ for FedAvg. All models are trained with the sgd optimizer for 150 rounds. The set up of the Celeba and Fake News data set are described in the supplementary material, as well as evaluation on the CIFAR-100 data set. Evasion attacks are $K=10$ step PGD attacks, with step size $\alpha_{\text{PGD}} = 0.01$ and a $l_2$ norm-ball perturbation budget of $\epsilon = 4.5$.

The effects of internal and external attacks are shown in Table \ref{tbl:cifar_fedavg_local}.
Given the data set $D_a$ and model $H_a$ of the adversary, and $D_v$ and $H_v$ of the victim client, the test accuracy (Test Acc.) is measured by the classification accuracy of data set $D_v$ by model $H_v$, and the robustness (Adv. Acc.) of models is measured by the classification accuracy of evasion attacks that are crafted from data $D_a$ of an adversarial client's model $H_a$ and evaluated by the victim model $H_v$. All participating clients perform internal evasion attacks to all other participating clients. Standard deviation of measurements are shown in parentheses, and measure the deviation between clients. External evasion attacks are performed by training a separate model with a different random seed with FedAvg, and using that model as the adversary $H_a$.

\begin{table}[t]
\centering
\caption{\small{Transferability of internal and external attacks for different training algorithms. Standard deviation in parentheses.}}
\begin{tabular}{l|lll}
\hline
CIFAR10   & Test Acc.    & \begin{tabular}[c]{@{}l@{}}Internal \\ Adv. Acc\end{tabular} & \begin{tabular}[c]{@{}l@{}}External \\ Adv. Acc\end{tabular} \\ \hline
FedAvg    & 0.93 (0.02)  & 0.00 (0.00)                                                  & 0.02 (0.02)                                                  \\
Local     & 0.38 (0.12)  & 0.33 (0.12)                                                  & 0.38 (0.12)                                                  \\
FedEM     & 0.76 (0.05)  & 0.13 (0.09)                                                  & 0.21 (0.05)                                                  \\ \hline
Celeba    & Test Acc.    & \begin{tabular}[c]{@{}l@{}}Internal \\ Adv. Acc\end{tabular} & \begin{tabular}[c]{@{}l@{}}External \\ Adv. Acc\end{tabular} \\ \hline
FedAvg    & 0.78 (0.09)  & 0.00 (0.00)                                                  & 0.13 (0.09)                                                  \\
Local     & 0.61 (0.24)  & 0.37 (0.23)                                                  & 0.34 (0.21)                                                  \\
FedEM     & 0.81 (0.18)  & 0.13 (0.20)                                                  & 0.26 (0.18)                                                  \\ \hline
Fake News & Test Acc.    & \begin{tabular}[c]{@{}l@{}}Internal \\ Adv. Acc\end{tabular} & \begin{tabular}[c]{@{}l@{}}External \\ Adv. Acc\end{tabular} \\ \hline
FedAvg    & 0.61 (0.32)  & 0.00 (0.00)                                                  & 0.25 (0.12)                                                  \\
Local     & 0.79 (0.31)  & 0.23 (0.34)                                                  & 0.48 (0.33)                                                  \\
FedEM     & 0.76 (0.29)  & 0.26 (0.30)                                                  & 0.48 (0.31)                                                  \\ \hline
\end{tabular}
\label{tbl:cifar_fedavg_local}
\end{table}

\begin{table}[t]
\centering
\caption{\small{Legitimate and adversarial inter-boundary distance ($I_d$), and gradient alignment measurements for the non-adversarially trained models for CIFAR-10 data set. Values in parentheses represent standard deviation across clients.}}
\begin{tabular}{l|lll}
\hline
CIFAR-10 & Leg. $I_d$  & Adv. $I_d$  & Grad. Align. \\ \hline
FedAvg   & 0.00 (0.00) & 0.00 (0.00) & 0.00 (0.00)  \\
Local    & 24.2 (11.8) & 24.0 (12.0) & 1.31 (0.53)  \\
FedEM    & 6.82 (4.18) & 4.48 (4.81) & 0.78 (0.47)  \\ \hline
\end{tabular}
\label{tbl:ib_benign}
\vspace{-3mm}
\end{table}

\emph{FedAvg displays some level of innate robustness against external but not internal evasion attacks}, with an external attack robustness of $0.13$ on Celeba, compared to a robustness of $0.00$ to internal attacks. 
When clients only train with their local data, the accuracies against external and internal evasion attacks are $0.38$ and $0.33$, respectively, for CIFAR-10. Robustness to evasion attacks is much higher for local training than federated learning for all data sets, but the test accuracy for benign inputs is much lower for local training, e.g., 0.38 compared to 0.93 of FedAvg on CIFAR10. Thus, the \emph{trade-off between accuracy and robustness against evasion attacks} is evident. Similar trade-offs can be seen across all data sets, except for the Fake News data set that has high test accuracy for local training. Personalized federated learning (FedEM) takes elements of both FedAvg and local training through personalization, and achieves relatively high test accuracy and robustness to internal and external attacks; it is analyzed more in depth in Section \ref{sec:pFedDef}.

Although theoretically characterizing the difference between models and its impact on evasion attack transferability is an open area of research, the relationship between the two have been analyzed with empirical methods. The \emph{inter-boundary distance} metric (IBD, elaborated in Supplementary material Section \ref{app:ib}) empirically measures the similarity of decision boundaries between two different models, with a lower measured value indicating more similar models \citep*{advspace}. 
The legitimate IBD measures the difference in models when classifying benign inputs, while the adversarial IBD measures the difference in models when classifying evasion attacks.
Table \ref{tbl:ib_benign} compares the average IBDs between clients of different distributed learning methods for CIFAR-10. Clients in a FedAvg system has no difference in boundary, while the distance subsequently increases for FedEM and local clients. A higher adversarial IBD is correlated with a lower rate of attack transferability. Similarly, the \emph{gradient alignment} metric compares the alignment of the input
gradients of the loss function computed for two models, and a lower value indicates higher alignment, leading to higher attack transferability \citep*{advtransfer}. The trends of gradient alignment in Table \ref{tbl:ib_benign} further follow the attack transferability rates and IBDs.

\section{FEDERATED ADVERSARIAL TRAINING} \label{sec:FAT}

The work presented in \citet*{FAT} is the first to consider utilizing adversarial training with federated learning. Adversarial training is a defense mechanism known to be reliable against perturbation attacks \citep*{adversarialEff}, especially against external evasion attacks for which perturbations are generated with respect to a substitute model. 
Intuitively, the goal of adversarial training is to introduce perturbed inputs into the training data set of a model, thus allowing the model to learn how to correctly classify perturbed inputs that it may later encounter.  In this section, we first describe the federated adversarial training process, and then empirically evaluate the robustness gained through adversarial training for FedAvg, local, and personalized federated learning against internal and external evasion attacks. 

\subsection{The Adversarial Training Process}

Formally, any defense against evasion
attacks can be formulated as a saddle point problem, with the goal of training a model that minimizes the empirical risk over a classification task, despite the adversary introducing input perturbations (i.e., using PGD as shown in Equation \ref{PGD}) that maximize the loss at each data point \citep*{madryAttacks}. 
The objective function of adversarial training is as follows:
\begin{align}\label{eq:pFAT}
    \min_{\theta} \mathbb{E}_{(x,y)\sim \Tilde{D}} \left[\max_{\delta \in S} L_{\Tilde{D}}(h_{\theta}, x+\delta,y)\right]
\end{align}
The perturbation $\delta$ added to the data is bounded within a budget $S$.
In words, we desire to find model parameters $\theta$ that minimize the expected maximum loss from perturbing the input $x$ by $\delta$.
When performing federated adversarial training, each client introduces adversarial examples into their training data set by using gradient information from their local model. The gradients are then aggregated centrally following the regular FedAvg protocol. 

\begin{table}[t]
\centering
\caption{\small{Transferability of attacks given different federated adversarial training algorithms. Standard deviation in parentheses.}}
\begin{tabular}{l|lll}
\hline
CIFAR10    & Test Acc.   & \begin{tabular}[c]{@{}l@{}}Internal \\ Adv. Acc\end{tabular} & \begin{tabular}[c]{@{}l@{}}External \\ Adv. Acc\end{tabular} \\ \hline
FAT        & 0.80 (0.05) & 0.30 (0.05)                                                  & 0.77 (0.06)                                                  \\
Local Adv. & 0.30 (0.09) & 0.28 (0.09)                                                  & 0.28 (0.09)                                                  \\
pFedDef    & 0.66 (0.07) & 0.48 (0.12)                                                  & 0.62 (0.07)                                                  \\ \hline
Celeba     & Test Acc.   & \begin{tabular}[c]{@{}l@{}}Internal \\ Adv. Acc\end{tabular} & \begin{tabular}[c]{@{}l@{}}External \\ Adv. Acc\end{tabular} \\ \hline
FAT        & 0.78 (0.08) & 0.18 (0.09)                                                  & 0.69 (0.10)                                                  \\
Local Adv. & 0.58 (0.24) & 0.35 (0.23)                                                  & 0.37 (0.23)                                                  \\
pFedDef    & 0.78 (0.22) & 0.35 (0.21)                                                  & 0.61 (0.21)                                                  \\ \hline
Fake News  & Test Acc.   & \begin{tabular}[c]{@{}l@{}}Internal \\ Adv. Acc\end{tabular} & \begin{tabular}[c]{@{}l@{}}External \\ Adv. Acc\end{tabular} \\ \hline
FAT        & 0.61 (0.32) & 0.00 (0.00)                                                  & 0.61 (0.32)                                                  \\
Local Adv. & 0.79 (0.30) & 0.48(0.32)                                                   & 0.40 (0.33)                                                  \\
pFedDef    & 0.79 (0.32) & 0.52 (0.32)                                                  & 0.53 (0.32)                                                  \\ \hline
\end{tabular}
\label{tbl:fat_attack}
\end{table}

\begin{table}[t]
\centering
\caption{\small{Inter-boundary distances ($I_d$), and gradient alignment measurements for the adversarially trained models for CIFAR-10 data set. Standard deviation across clients in parentheses.}}
\begin{tabular}{l|lll}
\hline
CIFAR-10   & Leg. $I_d$  & Adv. $I_d$  & Grad. Align. \\ \hline
FAT        & 0.00 (0.00) & 0.00 (0.00) & 0.00 (0.00)  \\
Local Adv. & 26.6 (13.6) & 33.3 (23.9) & 1.29 (0.53)  \\
pFedDef    & 8.81 (3.87) & 32.7 (18.6) & 0.85 (0.53)  \\ \hline
\end{tabular}
\label{tbl:ib_adv}
\vspace{-3 mm}
\end{table}

\subsection{Evaluation of Evasion Attacks Against Federated Adversarial Training}

We empirically characterize the effect of adversarial training for local and federated learning. The training set up and parameters are equivalent to that of the evaluation in Section \ref{sec:trans_eval}. For the adversarial training process, each client alters the input to a subset of its data ($G=0.5$) with adversarial perturbations generated by PGD. The adversarial perturbations are generated with $K=10$ steps, with a $l_2$ norm-ball perturbation budget of $\epsilon = 4.5$ and a step size of $\alpha_{\text{PGD}} = 0.01$. The results are shown in Table~\ref{tbl:fat_attack}.

Adversarial training significantly increases FedAvg's external attack robustness. For instance, FAT (FedAvg with adversarial training) achieves a robustness of 0.77 to external attacks for CIFAR-10, compared to a robustness of 0.02 for FedAvg (Table~\ref{tbl:cifar_fedavg_local}).
However, \emph{adversarial training alone does not provide sufficient robustness against internal attacks} in the federated learning setting, as adversarial perturbations by clients are created with respect to the gradient information of the already hardened models. FAT even achieves internal robustness of 0.00 on the Fake News data set.
The test accuracy and robustness of local training against internal and external attacks decrease for all datasets as a result of adversarial training; thus, \emph{the accuracy and robustness trade-off still exists}, as indicated by the still higher robustness of the local training methods despite the adversarial training of federated learning.
The training method pFedDef, which combines FedEM's personalized federated learning with adversarial training, does achieve higher internal robustness, comparable to that of local training. In Table \ref{tbl:ib_adv}, the adversarial inter-boundary distance as well as the gradient alignment metric increases for pFedDef compared to that of FedEM in Table \ref{tbl:ib_benign}, indicating that adversarial training combined with personalized learning pushes individual clients models further apart, contributing to the lower transferability of internal evasion attacks.
We next introduce pFedDef in Section~\ref{sec:pFedDef}.

\section{PERSONALIZED LEARNING WITH ADVERSARIAL TRAINING} \label{sec:pFedDef}

Given the limited benefits of federated adversarial training against internal attacks, as shown in Section~\ref{sec:FAT}, we aim to leverage the innate robustness against internal attacks demonstrated by local learning in Section~\ref{sec:transferability}.
Personalized federated learning, and its adversarial training counterpart pFedDef, transcends the accuracy to robustness trade-off by leveraging clients' non-i.i.d. distributions of data to create differences between client models (to reduce attack success) and have high test accuracy due to personalization of each model to client data \citep*{fedMTL21}.

\subsection{Personalized Federated Learning}

We formalize the objective of personalized federated learning by considering $C$ different clients. Each client has a different data distribution $D_c$, and it is desirable to fit a hypothesis, or model, $h_c$ for each client $c$. Given that $(x,y)\sim D_c$ represents the data drawn from $D_c$, letting $L_{D_c}$ denote the loss incurred by $h_c$ on $D_c$, we wish to solve:
\begin{align} \label{eq:FedEM}
\forall c \in [C], \min_{h_c \in H} \mathbb{E}_{(x,y)\sim D_c} L_{D_c}(h_c, x,y)
\end{align}
We assume that the clients solve this optimization problem by following the mixture-based framework proposed in \citet*{fedMTL21}, where each client's data distribution is modeled as a %weighted sum of a 
mixture of $M$ (unknown) underlying distributions $\Tilde{D}_m, \forall m \in [M]$. Each class label is probabilistically associated with each distribution. This framework subsumes many personalized federated learning approaches, including multi-task learning \citep*{fedMTL} and clustered federated learning \citep*{clustered_fl}. 
%\tdCarlee{Explain this statement more.}
We define a hypothesis $h_{\theta^*_m}$ for each underlying distribution $m$ that minimizes the expected loss for the classification task over that distribution, where $\theta \in \mathbb{R}^d$ are the model parameters. It is shown in Proposition 2.1 of~\citet*{fedMTL21} that the optimal hypothesis for a client is a linear sum of the hypotheses of each data distribution: $h^*_c = \sum_{m \in M} \pi^*_{c,m} h_{\theta^*_m}, \forall c\in[C]$. Here, $\pi^*_{c,m}$ represents the weight of the data drawn from distribution $m$ in client $c$'s mixture; generally, the $\pi^*_{c,m}$ are not known before training.

FedEM uses an expectation-maximization algorithm where clients individually solve for the weights  $\pi^*_{c,m}$, while jointly solving for the hypothesis for each distribution $h_{\theta^*_m}$. During the E-step, the probability that each data point is drawn from a certain distribution is updated using fixed values of $\pi_{c,m}$ and $h_{\theta_m}$. During the M-step, the distribution weights and hypothesis are updated using the probabilities computed in the E-step. Afterwards, the hypotheses are averaged at a central server and then returned to local clients.

\setlength{\algomargin}{0.8em}
\begin{algorithm}[t]
\SetAlgoLined
\textbf{Input:} 
Adv. Proportion $G$, Dataset Update Freq. $Q$, PGD steps $K$, Client resource $R_c$

\textbf{Collect Resource Information:} 
Quantify resource availability $R_c \in [0,1]$ indicating adversarial data proportion at each client $c\in[C]$. 

\textbf{For each training round} $t\in[T]$

\begin{enumerate}
    \item \textbf{Adversarial Data Set Update:} 
    If $t\%Q == 0$ and $t \neq 0$, update training adversarial training data set based on adversarial data set proportion $F_c \leq R_c$ calculated from the robustness propagation of Algorithm \ref{alg:robust_prop}.
    \item \textbf{FedEM:} Perform personalized federated learning using the adversarial data set developed by each of \\
    the clients. 
\end{enumerate}

\caption{pFedDef: Personalized Federated Adversarial Training}
\label{alg:pfeddef}
\end{algorithm}

\setlength{\algomargin}{0.8em}
\begin{algorithm}[t]
\SetAlgoLined
\textbf{Inputs:} 
Set adversarial data set proportion $G$, client adversarial data resource constraint $\{R_c\}_{c\in[C]}$. 

\textbf{Setting:} 
Set value $\Delta$ for increments of updating $F_c$ and number of steps for updating $I$.

\textbf{Consider Resource Constraints} 
for all clients $c\in[C]$,  if $G \leq R_c$, set $F_c \leftarrow G$, and if $G > R_c$, set $F_c \leftarrow R_c$

\textbf{Incremental Robustness Propagation} 
Heuristically minimize the objective function of Equation \ref{eq:opt} by running the following routine $I$ times:
\begin{enumerate}
    \item Randomly sample a client without replacement $c$ \\
    for which $F_c = G$ and $F_c \leq R_c$ (Clients with\\
    leftover resources)
    \item Incrementally increase $F_c \leftarrow F_c + \Delta$ until the objective function does not decrease, or the\\
    resource constraint for the sampled client is met.
\end{enumerate}

\caption{Robustness Propagation for pFedDef}
\label{alg:robust_prop}
\end{algorithm}

\subsection{Adversarial Personalized Federated Learning}

We next introduce pFedDef, a novel adversarial training algorithm for personalized federated learning. pFedDef leverages the differences between client models in the personalized federated learning setting to be robust to both external black-box and internal grey-box attacks in which adversary clients have limited knowledge of victim client models.
The pFedDef algorithm further accounts for federated learning clients' limited and diverse resources: 
some clients may not have the resources to generate adversarial data points. To compensate, clients with abundant resources can propagate adversarial learning to clients with fewer resources but similar data sets. Though similar \emph{robustness propagation} ideas are used by \citet*{bnprop} for FedAvg, we allow more fine-grained and dynamic client participation in adversarial training. Furthermore, pFedDef propagates robustness between underlying distribution hypotheses $h_m$, maintaining differences in models between clients. Although 
we use hypothesis weights $\pi_m$ unique to FedEM, any personalized federated learning method that can gauge similarities between client data distributions (e.g., through comparing clients' model gradients) can use a similar approach.

Algorithm \ref{alg:pfeddef} shows the pFedDef algorithm. 
A global desired adversarial data set proportion $G \in [0,1]$ is set, and once every $Q$ rounds, each client updates its local data set to include adversarial training points (generated using its current local model) based on $G$.
Some clients may not be able to generate adversarial training points that cover $G$ fraction of their data sets, due to local resource constraints $R_c \in [0,1]$. Thus, each client sets its true adversarial proportion $F_c \leq R_c$. The robustness propagation presented in Algorithm \ref{alg:robust_prop} is used to calculate $F_c$ for all clients, depending on their resource availability and underlying data distributions. Clients with more resources may set $F_c\geq G$, compensating for clients with limited resources. FedEM is then performed on top of the augmented data sets with the adversarial data points.

\textbf{Adversarial Robustness Propagation.} 
We formulate an optimization problem for adversarial robustness propagation, which attempts to achieve the desired adversarial data set proportion $G$ globally by inducing clients with ample resources to increase their local adversarial proportions $F_c$.
Formally, given $G$ and client resource constraints $R_c$, 
we desire to solve for $F_c$ at each client $c$ such that:
\begin{align} \label{eq:opt}
\min_{F\in [0,1]} \; &\sum_{m\in[M]} \left| \sum_{c\in[C]} (F_c |D_c| \pi_{c,m}) - G |D| \pi_{c,m} \right| \quad \\
{\rm s.t.} 
&F_c \leq R_c, \forall c \in [C] \nonumber
\end{align}
This objective aims to adversarially represent each of the hypotheses for underlying distributions in the FedEM framework proportionally to the prevalence of each distribution across clients. Algorithm \ref{alg:robust_prop} presents a heuristic solution to problem \ref{eq:opt}. In line 4.2, the adversarial proportion for client $c$ is incrementally increased by $\Delta$ until it either reaches the resource constraint or there is no reduction in problem \ref{eq:opt}. This algorithm can guarantee that the objective in Eq. (\ref{eq:opt}) monotonically decreases, and alternative algorithms with provable guarantees may be substituted.

\subsection{Evaluation of Personalized Federated Learning}

We empirically evaluate the test accuracy and robustness of FedEM. We assume $m = 3$ underlying distributions and set the other training and attack parameters as in the evaluations in Sections~\ref{sec:transferability} and~\ref{sec:FAT}. As seen in Table~\ref{tbl:cifar_fedavg_local}, FedEM achieves a high test accuracy of 0.76 for CIFAR-10. Although its innate internal attack robustness of 0.13 is lower than that of local training of 0.33 for CIFAR-10, FedEM displays higher robustness to internal and external attacks than FedAvg across all datasets.

As seen in Table~\ref{tbl:fat_attack}, \emph{pFedDef's 
internal attack robustness significantly outperforms that of FAT and is either better than or comparable to that of local adversarial training on all datasets}. Against external attacks, pFedDef maintains a high robustness, e.g., 0.62 for CIFAR-10, although lower than that of FAT (0.77).

\begin{figure}[t]
\centering
\includegraphics[width=2.2in]{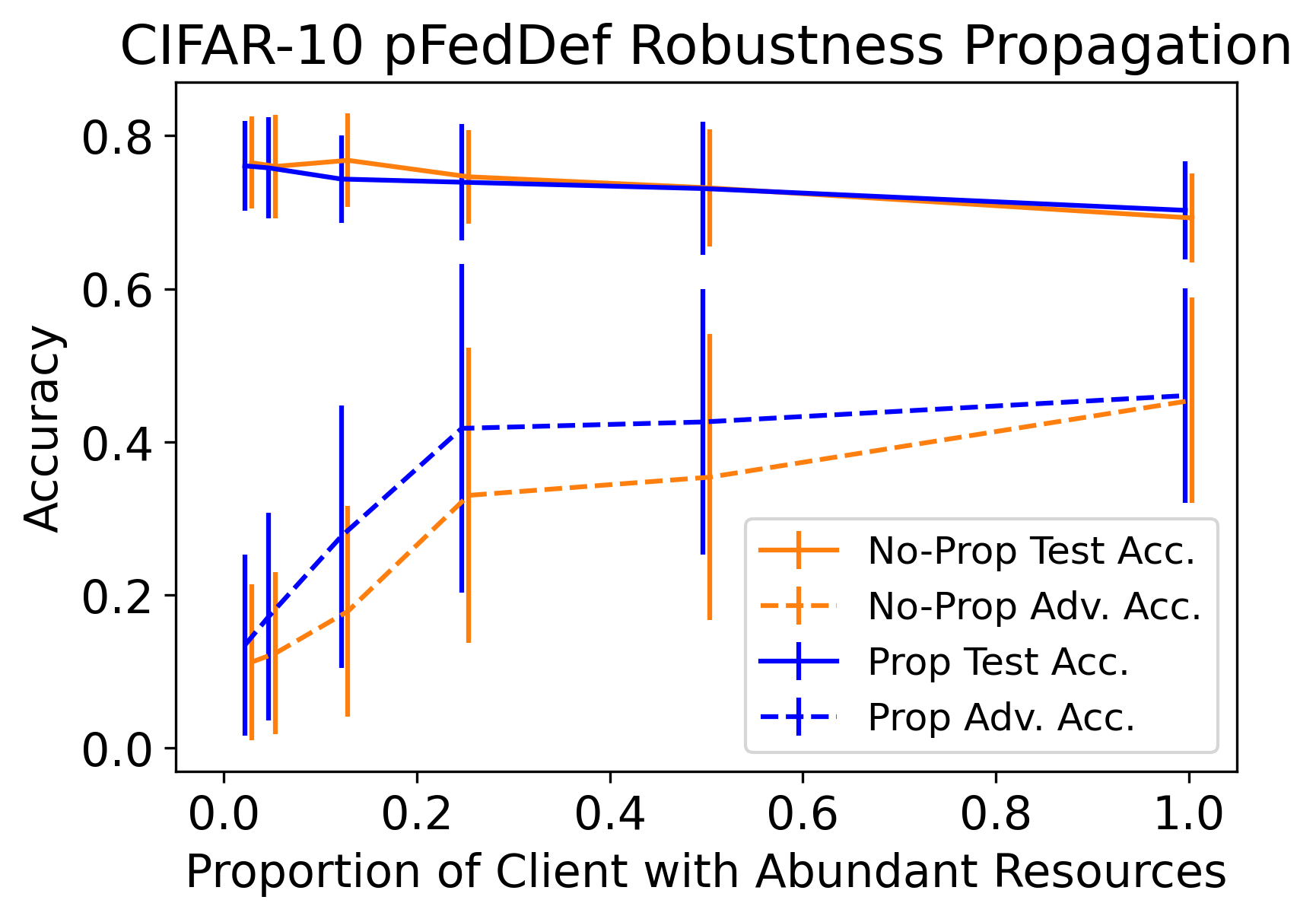}
\caption{\small{Performance of pFedDef with adversarial propagation, which gains higher robustness when resources are more limited.}}
\label{fig:cifar_prop}
\end{figure}

The effect of robustness propagation is examined in Figure~\ref{fig:cifar_prop} for CIFAR-10. The number of clients with ample resources for adversarial data set generation ($R_c = 0.7$) is gradually increased while that of clients with no resources ($R_c = 0$) is proportionally decreased. Test accuracy remains consistent regardless of resource availability and the presence of robustness propagation. Robustness consistently increases as the resources in the system increase in both cases. However, \emph{robustness propagation allows models to obtain higher robustness by leveraging resource availability at resource ample clients}, especially improving performance when overall system resources are low.

\begin{figure}[t]
\centering
\includegraphics[width=2.2in]{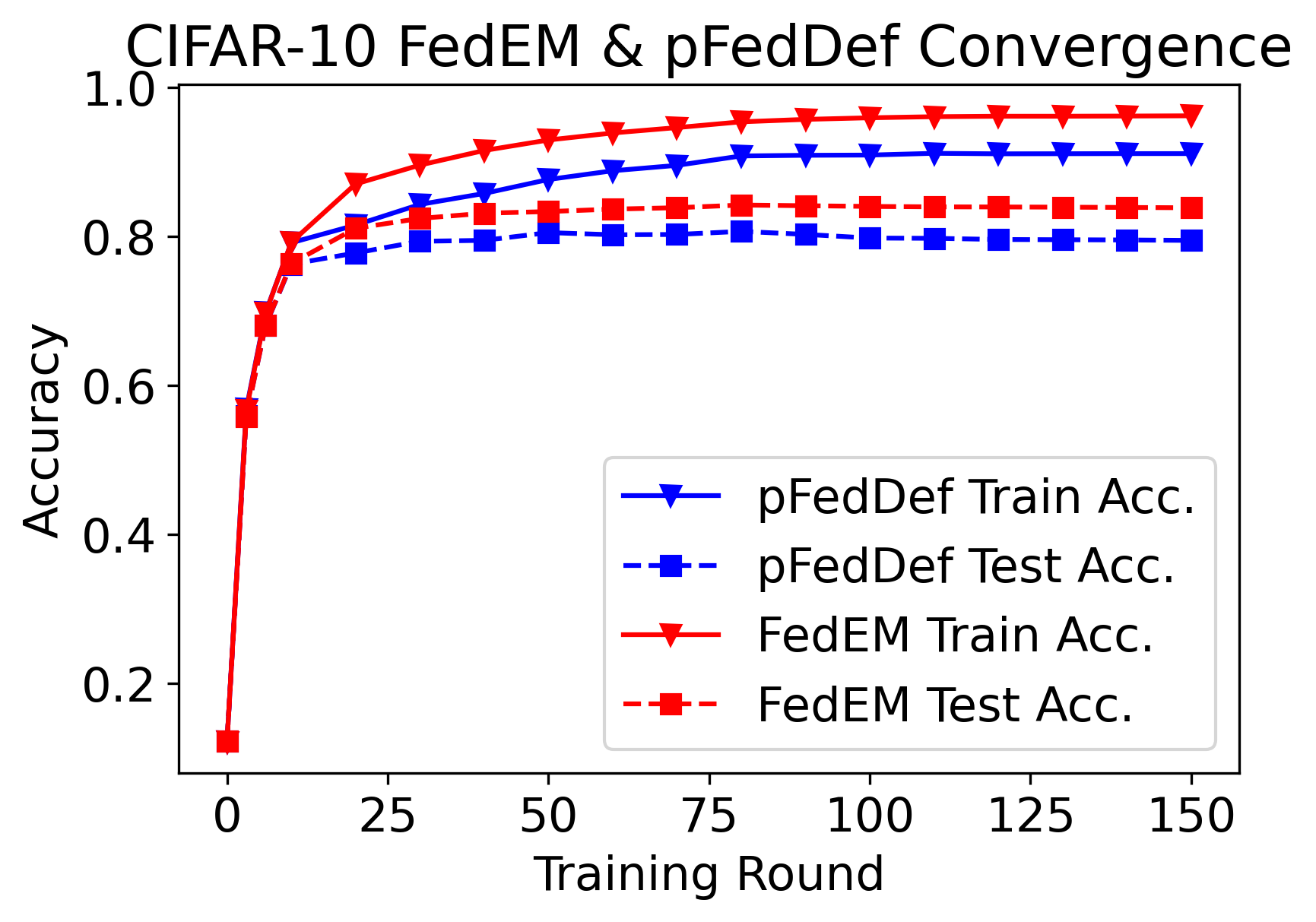}
\caption{\small{While both FedEM and pFedDef show model convergence, FedEM generally has higher accuracy than pFedDef. 
}}
\label{fig:converge}
\end{figure}

The convergence evaluation of pFedDef and FedEM for the CIFAR-10 data set is shown in Figure \ref{fig:converge}, 
where the convergence of pFedDef is expected to resemble that of the FedEM framework it is built upon (Theorem 3.2 of \citet*{fedMTL21}).
In both Figure \ref{fig:converge} and Tables \ref{tbl:cifar_fedavg_local} and \ref{tbl:fat_attack}, \emph{adversarial training generally slightly reduces test accuracy} as seen by comparing federated learning methods and their adversarially trained counterparts.
While the effect of adversarial training on model convergence is an open area of research, \citet*{ref:overparam_adv} indicate that adversarial training requires higher model capacity, leading to pFedDef suffering some accuracy loss. 
The use of high quality adversarial training points (e.g., multi-step PGD attacks) allows for convergence during adversarial training \citep*{ref:adv_converge}, which is aided by the resource management and robustness propagation of Algorithm \ref{alg:robust_prop}.

\begin{figure*}[t]
    \centering
    \begin{subfigure}[t]{0.33\textwidth}
    \includegraphics[width=0.9\textwidth]{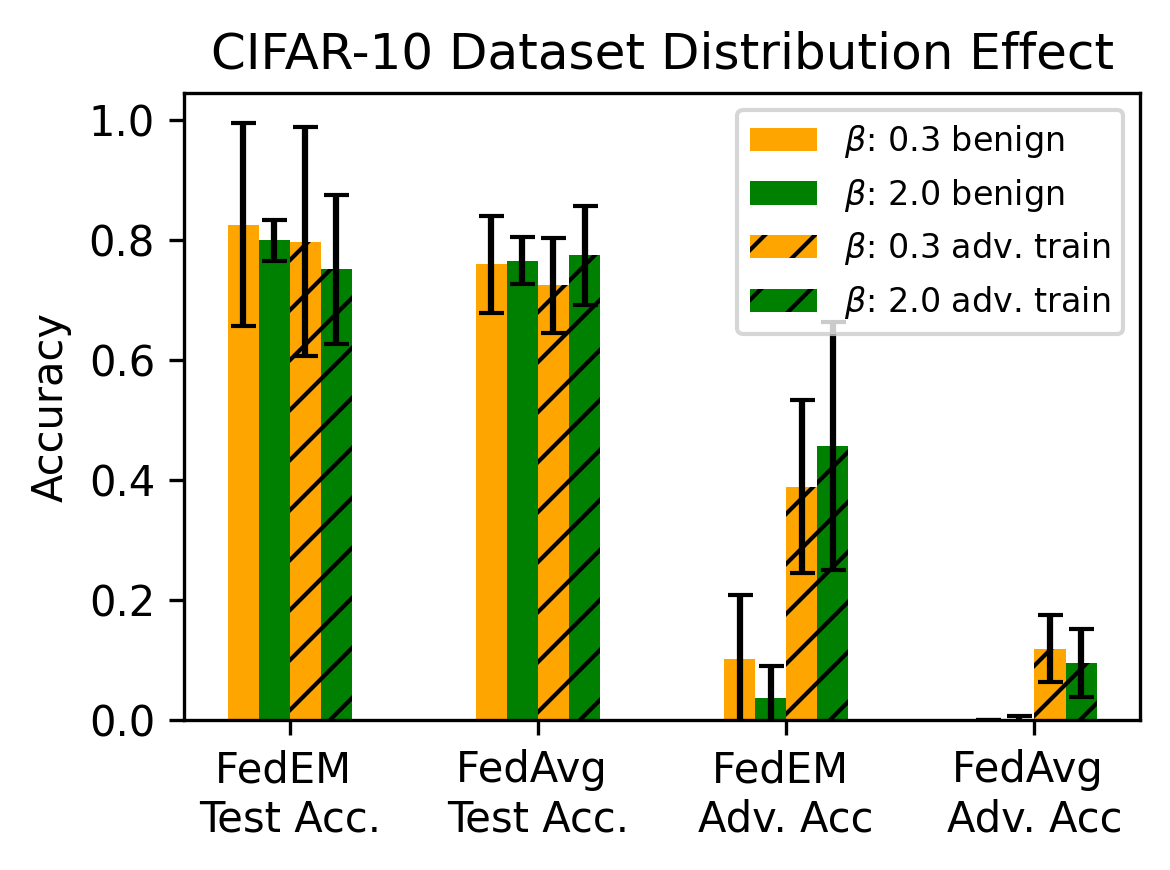}
    \caption{Data with higher value of $\beta$ is more i.i.d. than lower values, leading to lower robustness for FAT but higher for pFedDef.}
    \label{fig:cifar_alpha}
    \end{subfigure}
    \hspace{0.03cm}
    \begin{subfigure}[t]{0.32\textwidth}
    \includegraphics[width=0.92\textwidth]{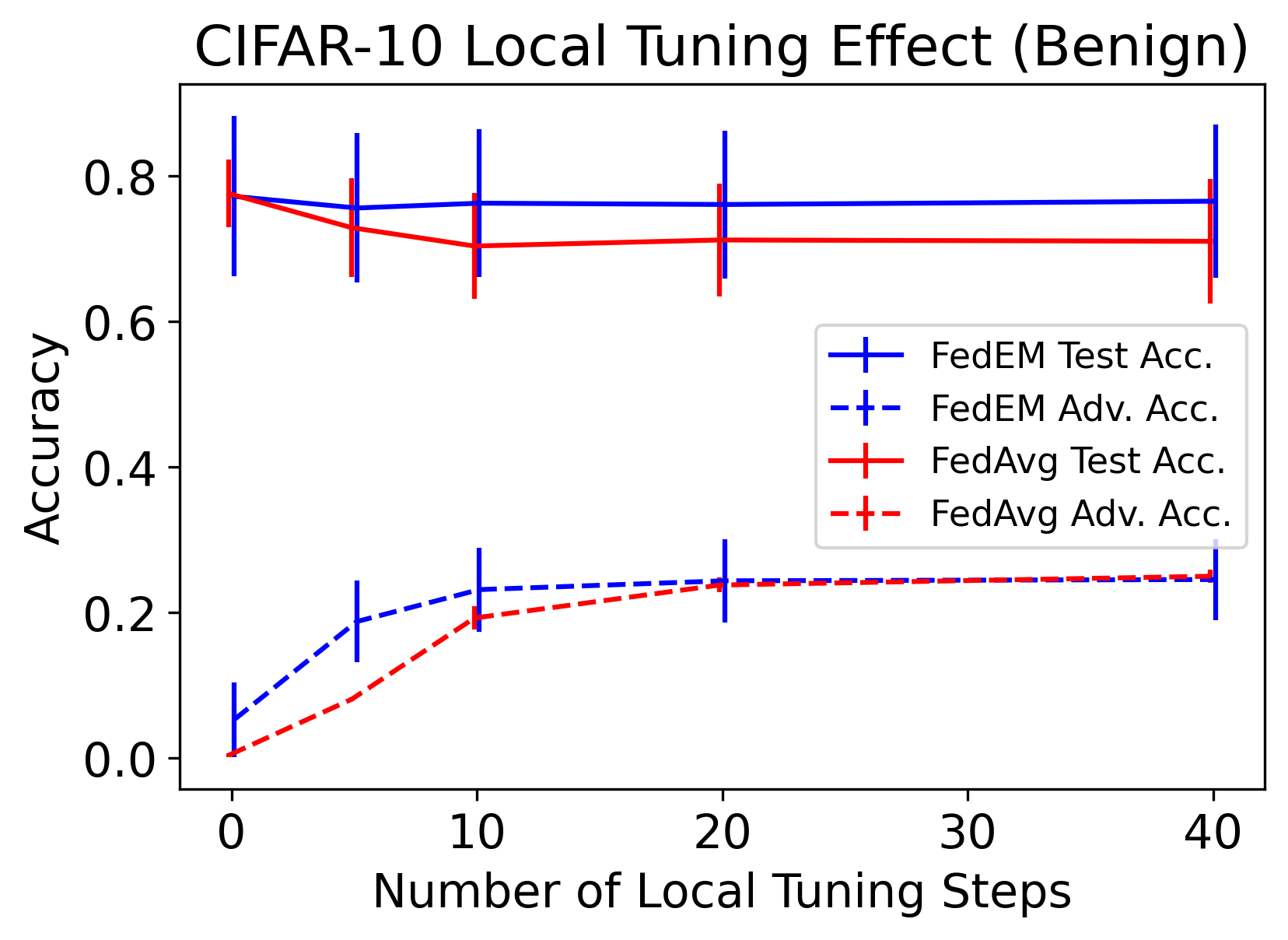}
    \caption{Additional local tuning increases the robustness of FedAvg and FedEM models.}
    \label{fig:cifar_tune_benign}
    \end{subfigure}
    \hspace{0.03cm}
    \begin{subfigure}[t]{0.32\textwidth}
    \includegraphics[width=0.92\textwidth]{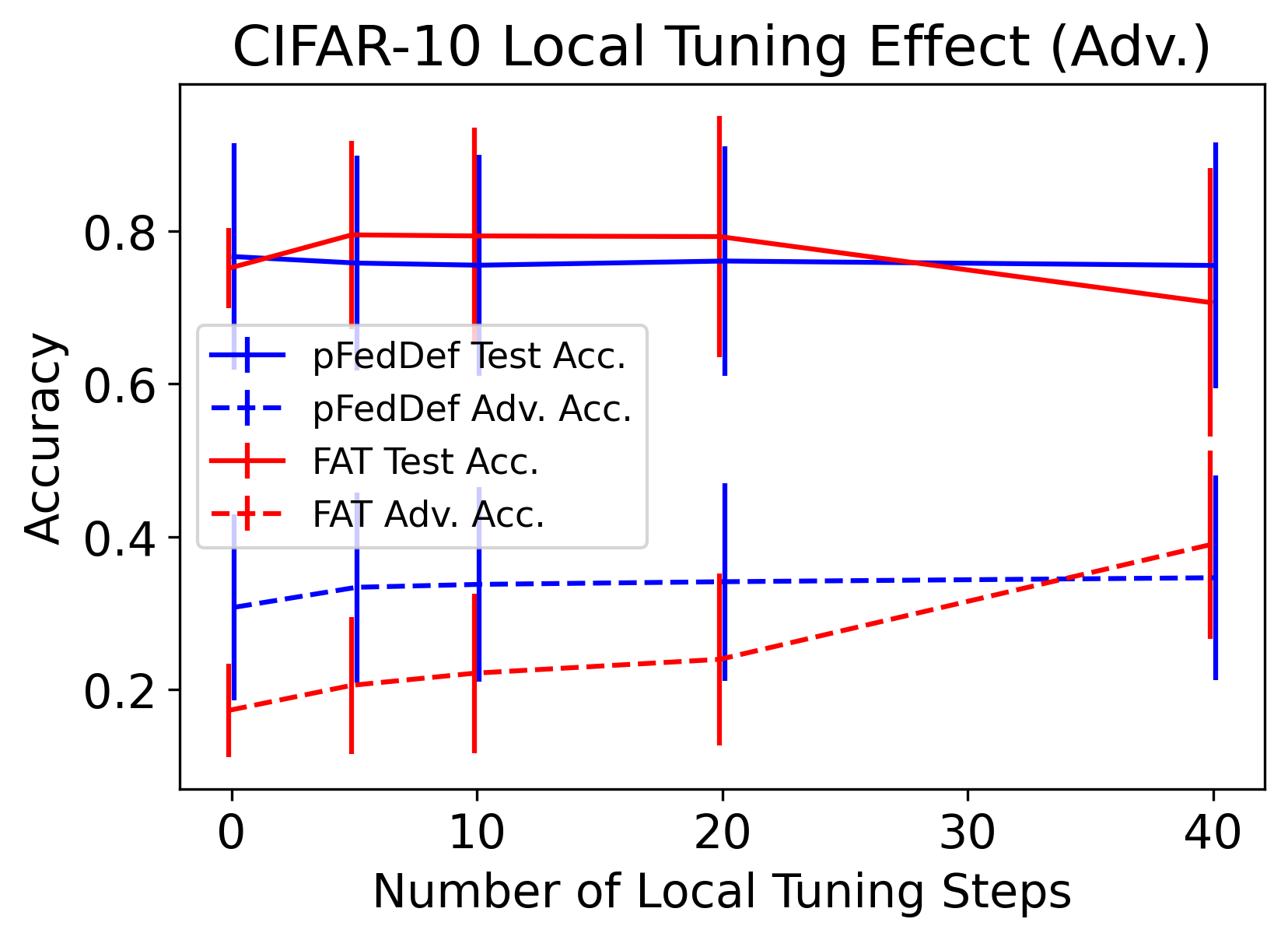}
    \caption{Additional local tuning increases the robustness of FAT models, and has a marginal effect on pFedDef models.}
    \label{fig:cifar_tune}
    \end{subfigure}
    \hspace{0.03cm}
  \caption{The pFedDef algorithm has consistently increased robustness given for varying data distributions. Local tuning allows FAT to gain similar levels of robustness to pFedDef.}
  \label{fig:group_plot_1}
\end{figure*}
\begin{figure*}[t]
    \centering
    \begin{subfigure}[t]{0.32\textwidth}
    \includegraphics[width=0.9\textwidth]{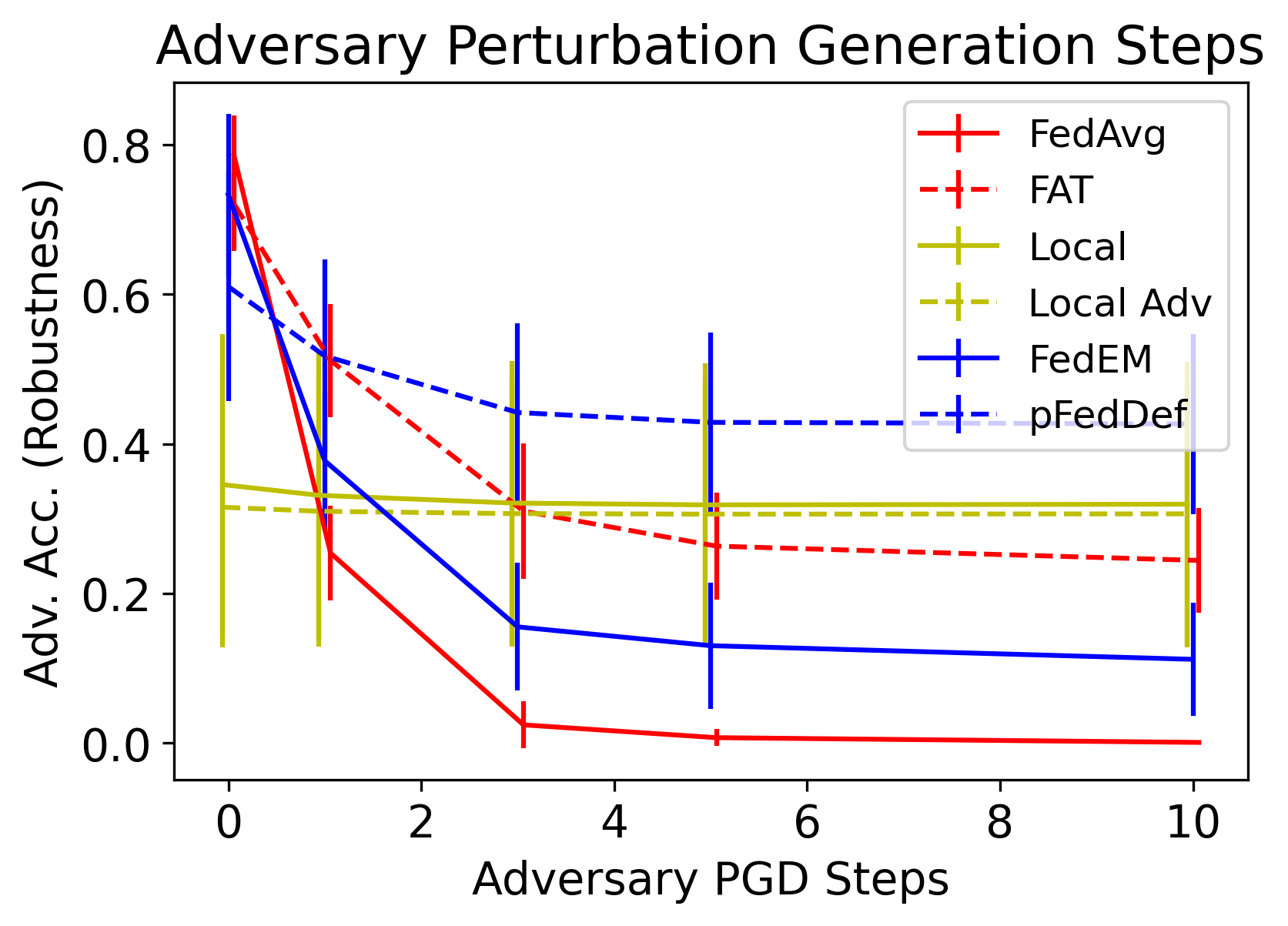}
    \caption{Internal evasion attacks with more attack steps have higher misclassification rates.}
    \label{fig:atk_step_sweep}
    \end{subfigure}
    \hspace{0.03cm}
    \begin{subfigure}[t]{0.32\textwidth}
    \includegraphics[width=0.9\textwidth]{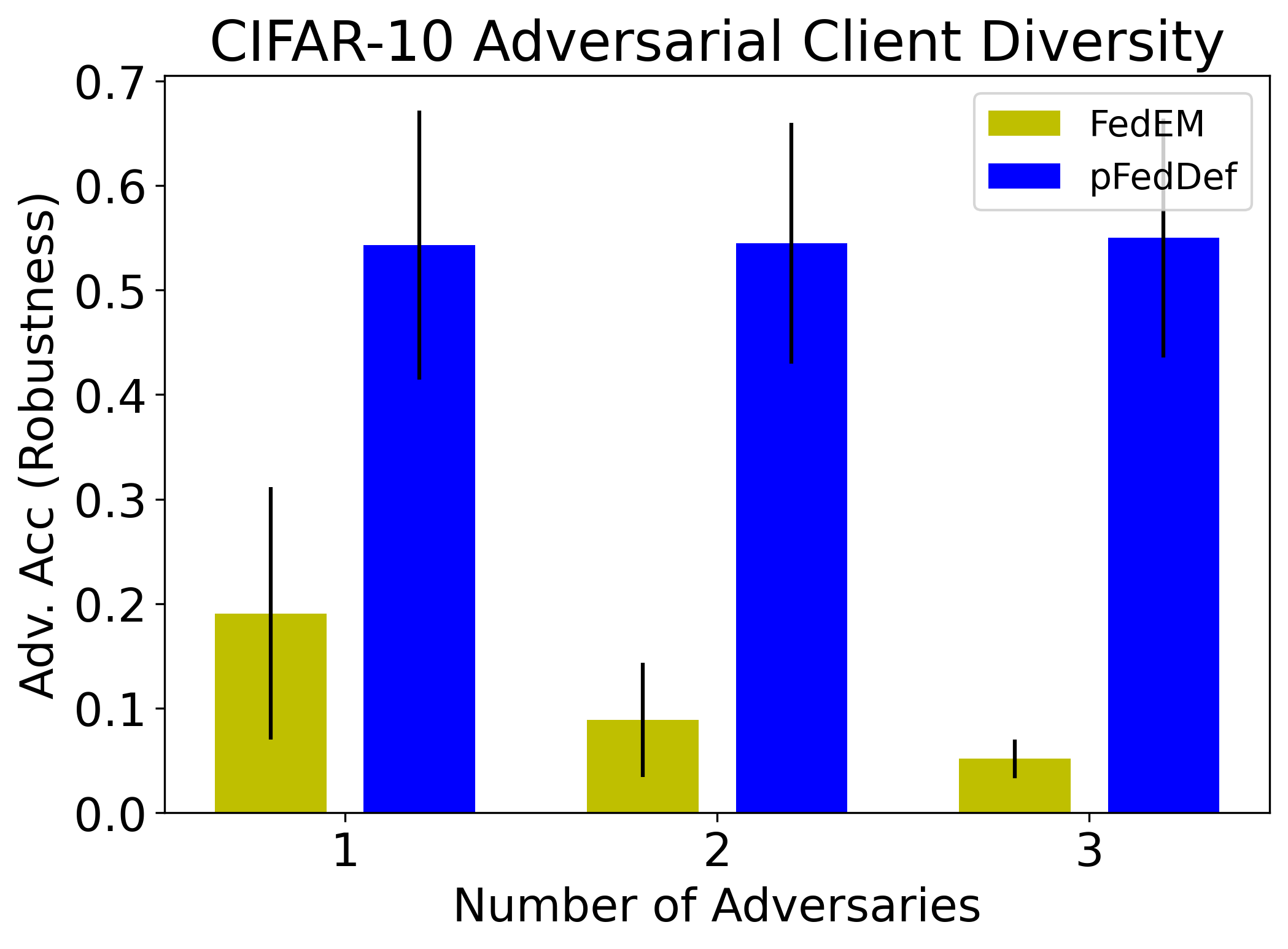}
    \caption{Ensemble attacks increase attack transferability in FedEM, but not in pFedDef.}
    \label{fig:cifar_ensemble}
    \end{subfigure}
    \hspace{0.03cm}
    \begin{subfigure}[t]{0.33\textwidth}
    \includegraphics[width=0.92\textwidth]{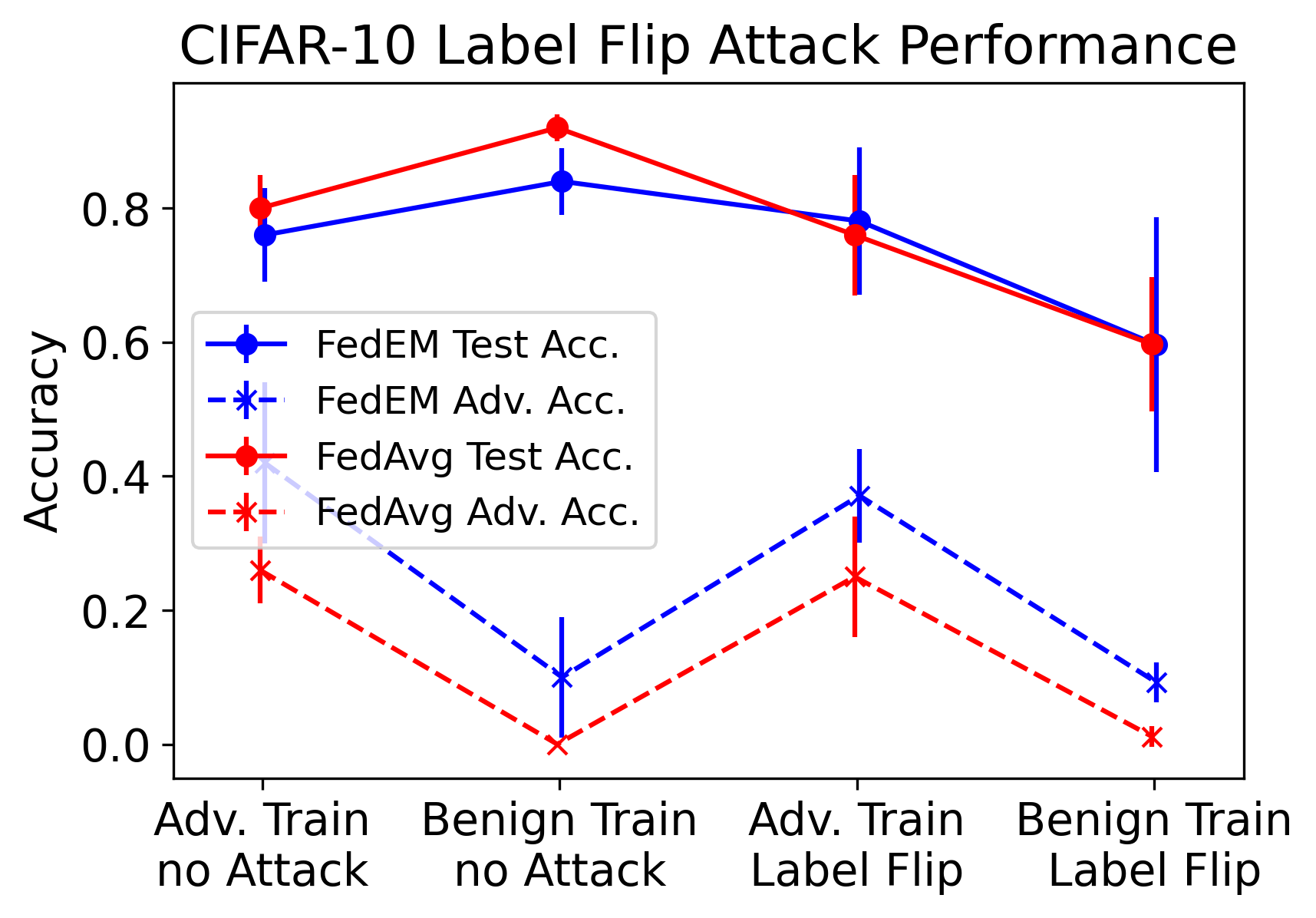}
    \caption{A Sybil label flipping attack is mitigated by FAT's and pFedDef's adversarial training.}
    \label{fig:cifar_sybil}
    \end{subfigure}
    \hspace{0.03cm}
  \caption{ The pFedDef algorithm maintains consistent performance subject to ensemble and label flipping attacks, while showing robustness against multi-step evasion attacks.}
  \label{fig:group_plot_2}
\end{figure*}
\vspace{-1mm}

\section{
EVALUATION: EXTENDED PERSONALIZATION AND ADVERSARIAL METHODS} \label{sec:exp}

In this section, we evaluate pFedDef beyond the results of Section~\ref{sec:transferability} and Section~\ref{sec:FAT}. We first examine the effectiveness of internal evasion attacks under different types of personalization across clients, looking at the effects of different data distribution settings and local tuning after federated learning. Afterwards, the effects of different adversarial models are examined beyond the scope of single client internal attacks, including the combination of internal evasion attacks with ensemble attack methods and Sybil attacks.

The parameters for training models and performing evasion attacks are identical to those found in Section~\ref{sec:transferability}, and adversarial training parameters are equivalent to that presented in Section~\ref{sec:FAT}. All measurements of robustness (Adv. Acc.) are with respect to internal evasion attacks. Experiments are performed on just the CIFAR-10 data set for brevity.

\subsection{Analysis of Personalization Methods}

Given the accuracy to robustness trade-off observed previously, we analyze how different data distributions at clients and local tuning to personalize federated learning models impact this trade-off.

\textbf{Data Set Distribution.} We vary the distribution of data across the clients with the parameter $\beta$ as in~\citet*{fedMTL21}; a low value of $\beta$ leads to a more non-i.i.d. split. A detailed explanation of this parameter and data splitting is available in the added supplementary material. In Figure \ref{fig:cifar_alpha}, the test accuracy and robustness of FedEM, FedAvg, and their adversarially trained counterparts (pFedDef and FAT, respectively) are analyzed for values $\beta=0.3$ and $2.0$.  As the data split becomes more i.i.d, the test accuracy of FedEM and pFedDef both fall, as the benefit of personalization is reduced on a more i.i.d. data split across clients. Conversely, the test accuracies of FedAvg and FAT increase with a more i.i.d. data split. 
The internal robustness for clients trained with FAT decreases as the data becomes more i.i.d., while pFedDef maintains higher robustness than FAT in both data split settings. The robustness of FedEM decreases as the data becomes more i.i.d., while it increases for pFedDef.

\textbf{Local Tuning.}
After federated learning has taken place, clients may perform local tuning by further training their model with local data sets, further differentiating their models. 
Local tuning acts as the most basic method of personalization between clients, and can potentially serve as a baseline for comparison or a substitute for FedEM when implementing pFedDef. The effects of local tuning on FedAvg and FedEM are presented in Figure \ref{fig:cifar_tune_benign}, while the same for FAT and pFedDef are presented in \ref{fig:cifar_tune}.
The effects of local tuning on FedAvg and FAT reflect the accuracy to robustness trade-off --
as more rounds of local tuning take place, the test accuracies for FAT and FedAvg eventually decrease due to over-fitting, while the robustness for FAT and FedAvg consistently increase, as the models of different clients diverge.
The effects of local tuning on the test accuracy are less pronounced for FedEM and pFedDef trained models, as clients already have personalized models. 
While FedEM gains some robustness with local tuning, pFedDef maintains similar levels of robustness before and after local tuning.

\subsection{Evaluation of Extended Adversarial Threats}

We evaluate different adversaries beyond a single malicious client, including multiple colluding adversarial clients performing ensemble attacks and train time (Sybil) attacks.

\textbf{Multi-Step Evasion Attacks.} In Figure~\ref{fig:atk_step_sweep}, we analyze FedAvg's, local training's, and FedEM's robustness against internal evasion attacks with and without adversarial training, given a different number of PGD-steps used by the adversary to craft the attacks. As discussed by \citet*{REN2020346}, evasion attacks crafted with a multi-step procedure are often more potent than single-step attacks, and may successfully deceive models developed with adversarial training. Against FedAvg and FedEM, internal evasion attacks lower the classification accuracy as the number of attack steps are increased. FAT (FedAvg + adversarial training) and pFedDef (FedEM + adversarial training) experience lower accuracy losses, with pFedDef experiencing the least severe accuracy drop, highlighting the effectiveness of personalization against internal evasion attacks. Local training and adversarial local training are largely unaffected by the increase in attack steps due to a higher difference in models between clients, as well as a lower general test accuracy.

\textbf{Ensemble Attacks.} As shown by \citet*{bagboostensemble}, generating \emph{ensemble} evasion attacks with gradient information from multiple models increases the success rate of attacks. Ensemble evasion attacks act as the inverse of personalization: more ensemble members with unique models increase the space covered by the adversary, and thus the amount of information the adversary knows about the clients.
We examine the transferability of ensemble attacks for FedEM and pFedDef by assuming a varying number of cooperating adversarial clients. We generate evasion attacks with different combinations of 3 adversarial clients, each associated with data drawn from 
one of the $m\in[M=3]$ global data distributions.
Here, the adversarial perturbations generated for the same input from different clients are averaged to perform an ensemble attack. For FedEM, when the perturbations are generated from more clients, the attack success rate increases, as seen in Figure \ref{fig:cifar_ensemble}. However, pFedDef's adversarial training manages to defend the entire hypothesis space, and dramatically improves robustness even against ensemble attacks with many adversaries. Thus, even if an adversary generates attacks with various configuration of weights $\pi_{c,m}$ under the FedEM model, pFedDef is still robust against such ensemble attacks

\textbf{Sybil Attacks.} We next evaluate FedEM's, FedAvg's, FAT's, and pFedDef's robustness to a Sybil data poisoning attack. Here, 10 of the 40 clients in the system perform a label flipping attack by scrambling the labels in their training sets during the training phase to reduce the performance of the models of other clients \citep*{trimmedmean}. As seen in Figure \ref{fig:cifar_sybil}, this attack reduces the test accuracy for the non-adversarially trained models of FedEM and FedAvg. However, although FAT and pFedDef are designed to increase robustness against evasion attacks at test time, they also provide robustness against these training-time Sybil attacks: their adversarial training provides more exposure to perturbed and abnormal data points.
While byzantine robust aggregation methods can also defend against such Sybil attacks for FAT and FedAvg \citep*{Krum}, FedEM and pFedDef  are not necessarily compatible with such methods as model aggregation occurs for multiple hypotheses. 
Without such aggregation, FedEM and pFedDef may be frail against scaling attacks, where adversaries upload high magnitude noise for aggregation \citep*{diffprivacy}.

\section{CONCLUSION}
\label{sec:conc}

While the use of adversarial training in the context of federated learning has been explored in recent years, such defenses do not explicitly consider attacks crafted by adversarial clients participating in the federated learning system. Depending on the federated learning method employed, such adversarial clients possess grey- or white-box information of the learned model and generate effective internal evasion attacks. In this paper, we characterize the impact of these internal evasion attacks on different forms of federated learning and then introduce pFedDef, an adversarial training framework built upon \emph{personalized} federated learning in order to reduce the model information shared between participating clients. Compared to existing federated adversarial training methods, pFedDef significantly increases robustness to internal transfer attacks while maintaining robustness against external black-box attacks. 
Moving forward, we can examine the use of smart local tuning methods to reduce attack transfer rates between federated clients with similar data distributions, as well as finding defenses for personalized federated learning systems against Sybil attacks, e.g., new byzantine robust aggregation methods.

\newpage

\subsubsection*{Acknowledgments}
\vspace{-1mm}
This research was partially supported by the 
Nicholas Minnici (E '59) Dean's Graduate Fellowship in Electrical and Computer Engineering and by the
CyLab Security and Privacy Institute, both in affiliation with Carnegie Mellon University.
Special thanks to Yixin Yang of Carnegie Mellon University (ECE undergraduate, class of 2024) who has helped compile and organize the code.

\bibliography{references}
% \bibliographystyle{unsrt}

% \clearpage
\clearpage
\appendix

\thispagestyle{empty}
\onecolumn 
% \makesupplementtitle

\renewcommand{\theHsection}{arabicsection.\thesection}
\renewcommand\thesection{\Alph{section}}

\section*{SUPPLEMENTARY MATERIAL}
\setcounter{section}{0}

\section{TABLE OF VARIABLES}

\begin{table}[h]
\centering
\caption{\small{Variables and notation used for transferability characterization and the pFedDef algorithm.
The data distribution parameter $\beta$ is explained in supplementary material Section \ref{sec:data-split}.}}
\begin{tabular}{l|l|l|l}
\hline
$c \in [C]$     & Clients of distributed system.                                                                                                                   & $I_d(h, h',x)$           & \begin{tabular}[c]{@{}l@{}}Inter-boundary distance between \\ models $h$ and $h'$ based on point $x$\end{tabular}             \\ \hline
$m \in [M]$     & \begin{tabular}[c]{@{}l@{}}Number of underlying data\\ distributions assumed to exist\end{tabular}                                               & $G$                      & \begin{tabular}[c]{@{}l@{}}Desired adversarial training proportion \\ for  pFedDef\end{tabular}                               \\ \hline
$(x,y)\sim D_m$ & \begin{tabular}[c]{@{}l@{}}Data drawn from underlying data set\\  $m\in[M]$\end{tabular}                                                         & $F_c \leq R_c$           & \begin{tabular}[c]{@{}l@{}}Actual adversarial proportion for pFedDef \\ for client $c$ bounded by resource $R_c$\end{tabular} \\ \hline
$h_m \in [H]$   & \begin{tabular}[c]{@{}l@{}}Hypothesis to fit data of underlying \\ distribution  $m\in[M]$\end{tabular}                                          & $Q$                      & \begin{tabular}[c]{@{}l@{}}Number of rounds between generating new \\ adversarial training data in pFedDef\end{tabular}       \\ \hline
$\pi_{c,m}$     & \begin{tabular}[c]{@{}l@{}}Proportion of underlying distribution \\ $m$ at client $c$ ($\forall c, \sum_{m \in [M]} \pi_{c,m} = 1$)\end{tabular} & $K$                      & Number of steps in multi-step PGD attack                                                                                      \\ \hline
$d(h, x)$       & \begin{tabular}[c]{@{}l@{}}Direction between $x$ and a different \\ label point $x'$  for model $h$\end{tabular}                                 & $\delta \in S$           & \begin{tabular}[c]{@{}l@{}}Perturbation $\delta$ added to data point $x$ \\ bounded by perturbation budget $S$\end{tabular}   \\ \hline
$N_d(h, x)$     & \begin{tabular}[c]{@{}l@{}}Distance from $x$ to decision boundary \\ for model $h$\end{tabular}                                                  & $\alpha$, $\alpha_{PGD}$ & \begin{tabular}[c]{@{}l@{}}Federated learning learning rate and PGD \\ step-size parameter\end{tabular}                       \\ \hline
\end{tabular}
\end{table}

\section{RESOURCES AND ASSETS} \label{app:ext_rsrc}

The code used to run the experiments is based on the pFedDef library that is available at \url{https://github.com/tj-kim/pFedDef_v1}. An introduction to and instructions on how to use the pFedDef library has been published in the Software Impacts Journal \citep*{software_impacts}.
pFedDef is based on the FedEM implementation created by  \citet*{fedMTL21}, and the code used to run their experiments is found at \url{https://github.com/omarfoq/FedEM}. Our work has adjusted the work presented in FedEM with the following changes:
\begin{itemize}
    \item We introduce the adversarial training mechanisms for different types of distributed learning, including pFedDef that utilizes adversarial robustness propagation.
    \item We build an transfer attack analyzer that can perform and analyze internal transfer attacks between federated clients. This includes the inter-boundary distance measurement tools between clients \citep*{advspace}.
    \item We add an ensemble attack mechanism and label flip attack mechanism.
\end{itemize}

\section{DATA SETS EXPLANATION}\label{app:data_exp}

\textbf{CIFAR.} The CIFAR-10 and CIFAR-100 data sets are selected to analyze the trends of pFedDef for two similar classification tasks of varying sizes. The CIFAR-100 model is also trained on MobileNetV2. The training and attack parameters of CIFAR-100 are equivalent to that of CIFAR-10, except that the number of clients in the system is 50, and the adversarial proportion is set at $G=0.5$. The data is artificially split between clients in a non-i.i.d. manner for both data sets with $\beta = 0.4$. As seen in Table \ref{tbl:cifar100}, CIFAR-100 displays the accuracy to robustness trade-off seen in Table \ref{tbl:cifar_fedavg_local} and \ref{tbl:fat_attack} for CIFAR-10. 

\begin{table}[h]
\centering
\caption{\small{Transferability of attacks for the CIFAR-100 data set given different federated learning methods and adversarial training counterparts. Values in parentheses represent standard deviation across clients.}}
\begin{tabular}{l|lll}
\hline
\begin{tabular}[c]{@{}l@{}}CIFAR 100\\ \end{tabular} & Test Acc.   & \begin{tabular}[c]{@{}l@{}}Internal \\ Adv. Acc.\end{tabular} & \begin{tabular}[c]{@{}l@{}}External\\ Adv. Acc.\end{tabular} \\ \hline
FedAvg                                                   & 0.31 (0.06) & 0.01 (0.00)                                                   & 0.02 (0.01)                                                  \\
Local                                                    & 0.29 (0.07) & 0.04 (0.03)                                                   & 0.02 (0.05)                                                  \\
FedEM                                                    & 0.34 (0.10) & 0.03 (0.01)                                                   & 0.02 (0.02)                                                  \\ \hline
FAT                                                      & 0.37 (0.07) & 0.07 (0.01)                                                   & 0.31 (0.06)                                                  \\
Local Adv.                                               & 0.26 (0.07) & 0.07 (0.04)                                                   & 0.09 (0.06)                                                  \\
pFedDef                                                  & 0.35 (0.09) & 0.16 (0.06)                                                   & 0.29 (0.05)                                                  \\ \hline
\end{tabular}
\label{tbl:cifar100}
\end{table}

\textbf{Celeba.} The Celeba data set is a large-scale data set with celebrity images, each with 40 binary labels, from LEAF, a bench-marking framework for federated learning \citep*{leaf}. This data set is selected for analysis as the distribution of data across clients follows a more realistic pattern than artificial division of data amongst clients used for other data sets. We combine 4 binary classification tasks (Smiling, Male, Eyeglasses, Wearing Hat) to formulate a classification problem with 16 classes. The images are reshaped to 50x50 shaped tensors. The Celeba model is  trained on MobileNetV2, with equal training and attack parameters to CIFAR-10 except the number of training rounds is set at 100. The results for Celeba in Table \ref{tbl:cifar_fedavg_local} and \ref{tbl:fat_attack} display similar patterns to other data sets analyzed.

\textbf{Fake News.} The FakeNews data set is a multi-dimensional data repository for various fake news detection related researches \citep*{fake_news_set}. In order to perform evasion attacks on fake news classification tasks, we extract the textual features from the news article using BERT \citep*{devlin2018bert} and then use [CLS] embedding of the news article as input to the MLP classification head. For this system, the attacker's motivation is to falsify the data such that a fake news article is classified as legitimate. The results for the FakeNews data set in Table \ref{tbl:cifar_fedavg_local} and \ref{tbl:fat_attack} display similar patterns to other data sets analyzed.

\begin{figure}[h]
\begin{minipage}[t]{0.475\columnwidth}
\centering
  \includegraphics[width=0.7\linewidth]{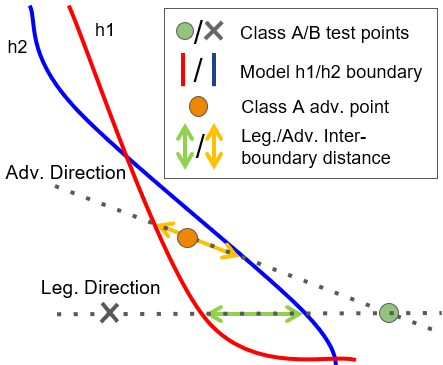}
  \caption{\small{Inter-boundary distance between models $h_1$ and $h_2$.}}
  \label{fig:ib_dist}
\end{minipage}\hfill % maximize horizontal separation
\begin{minipage}[t]{0.475\columnwidth}
\centering
  \includegraphics[width=0.8\linewidth]{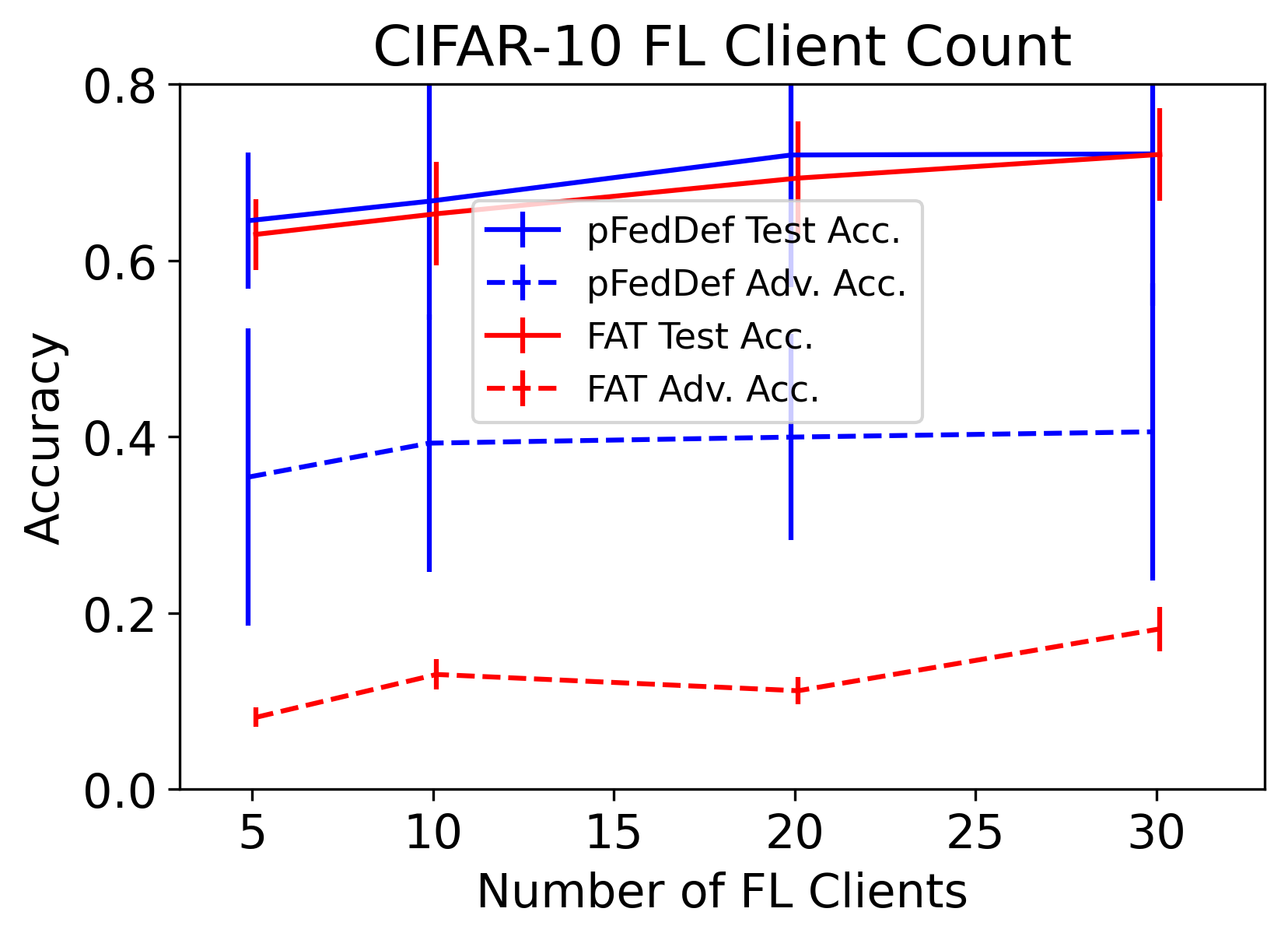}
  \caption{\small{Number of clients participating in federated learning increased shows small increase in test accuracy and robustness.}}
  \label{fig:client_count}
\end{minipage}
\end{figure}

\section{
BACKGROUND KNOWLEDGE: INTER-BOUNDARY DISTANCE} \label{app:ib}

The metric of \emph{inter-boundary distance} introduced by \cite{advspace} measures the distance between the decision boundaries of two models. The unit-norm direction vector between any point $x$ that is classified correctly by the two models in comparison, and the closest point in the $\ell_2$ distance $x'$ misclassified by $h_c$ is $d(h, x) := \frac{x'-x}{|| x' - x||_2}$.
Given model $h_{c}$ for client $c$, the \emph{legitimate direction} $d_{\text{leg}}(h_{c},x)$ is defined for each data point $x$ and the closest data point $x'$ with a different class label from $x$. The \emph{adversarial direction} $d_{\text{adv}}(h_c,x)$
is similarly defined by $x$ and an adversarial example $x' = x+\delta$ that is misclassified by model $h_c$. Given a direction $d$ (e.g., $d_\text{leg}$ or $d_\text{adv}$), the minimum distance $N_d$ between point $x$ to the decision boundary of model $h_c$ is:
\begin{align}\label{eq:mindist}
    \text{N}_d(h_{c},x) := \min \epsilon \\
    \text{s.t. } h_c(x + \epsilon \cdot d) \neq h_c(x), \epsilon > 0 \nonumber
\end{align}
Given a point $x$ and a direction $d$ computed according to a model $h_c$, the inter-boundary distance $I_d$ between two different models $h_{c}$ and $h_{c'}$ is defined as:
\begin{align}\label{eq:interdist}
    \text{I}_d(h_{c},h_{c'},x) := |\text{N}_d(h_{c},x) - \text{N}_d(h_{c'},x) |
\end{align}

Smaller inter-boundary distances indicate more similar models, inducing high transferability of attacks from one model to another. A visual depiction of the inter-boundary distance between two models in the legitimate and adversarial directions is shown in Figure \ref{fig:ib_dist}. We use the inter-decision boundary metric to quantify the similarity of clients' models in the FedAvg, FedEM, and local training settings, and compare it to empirical transferability values for CIFAR-10 in Table~\ref{tbl:ib_benign} and Table \ref{tbl:ib_adv}. Small inter-boundary distances for the legitimate direction indicate similarity in classification tasks for models, while small distances for the adversarial direction indicate the potential for attack transferability between models. The inter-boundary distance is the highest for local training for both regular and adversarial training, followed by FedEM and pFedDef. Overall, adversarial training increases the inter-boundary distance, especially the adversarial inter-boundary distance (Adv. $I_d$) for both local training and FedEM. 

\section{EXTENDED EVALUATION}

\begin{figure*}[t]
    \centering
    \begin{subfigure}[t]{0.32\textwidth}
    \includegraphics[width=1.0\textwidth]{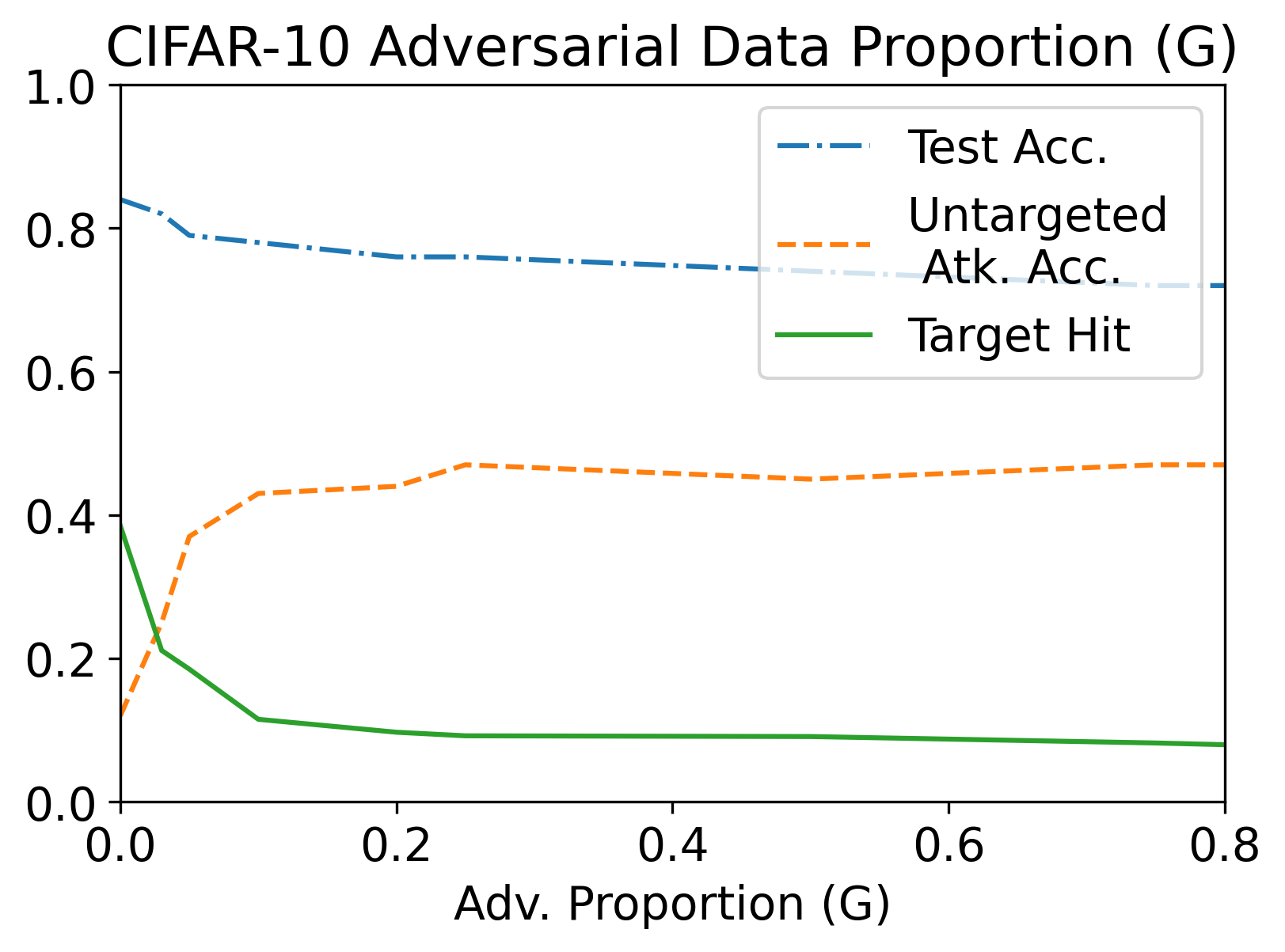}
    \caption{High pFedDef adversarial proportion increases robustness with diminishing returns.}
    \label{fig:G}
    \end{subfigure}
    \hspace{0.03cm}
    \begin{subfigure}[t]{0.32\textwidth}
    \includegraphics[width=1.0\textwidth]{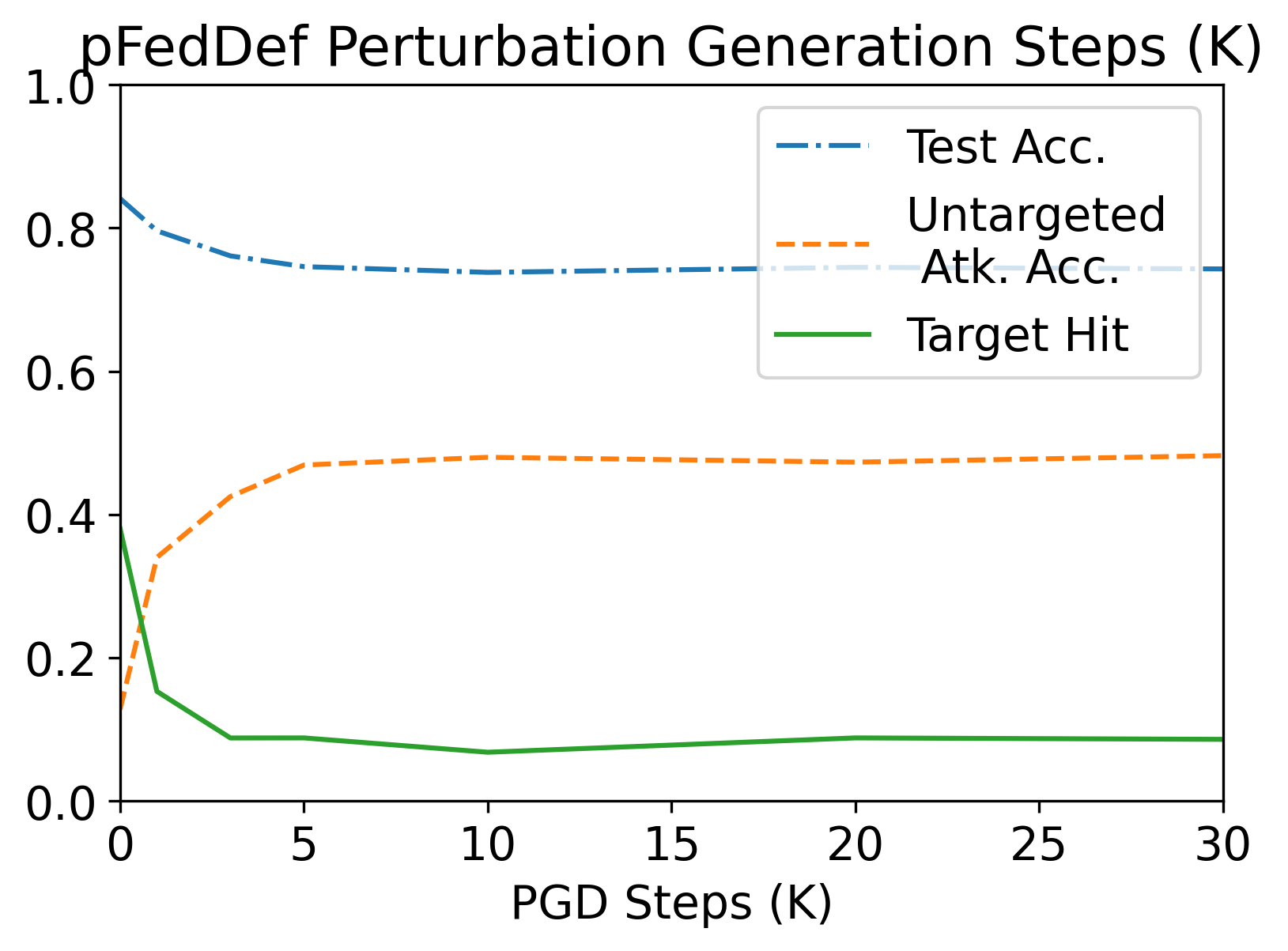}
    \caption{More PGD steps during pFedDef increases robustness with diminishing returns.}
    \label{fig:K}
    \end{subfigure}
    \hspace{0.03cm}
    \begin{subfigure}[t]{0.32\textwidth}
    \includegraphics[width=1.0\textwidth]{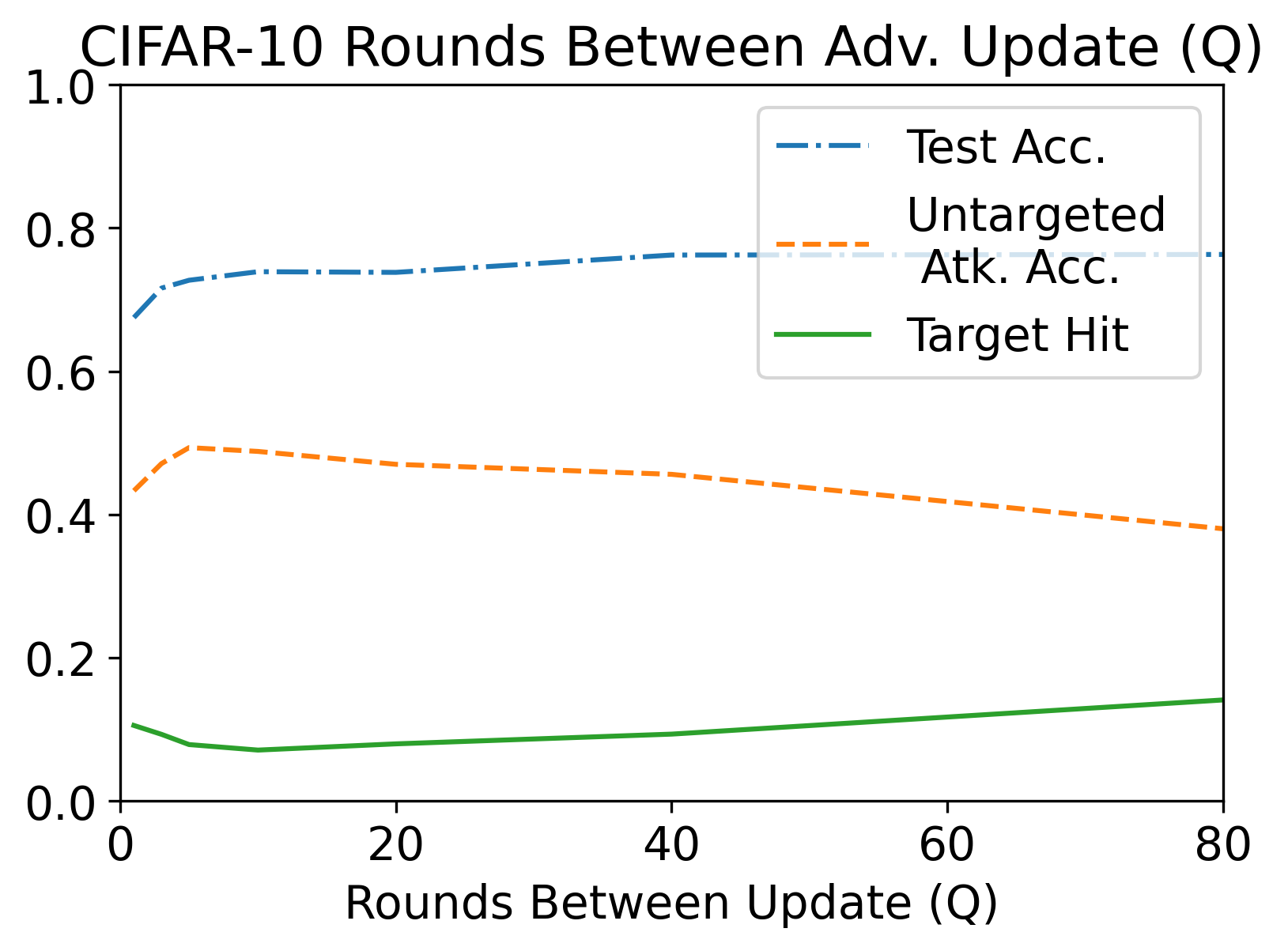}
    \caption{More rounds between adv. data set updates overall reduces robustness.}
    \label{fig:Q}
    \end{subfigure}
    \hspace{0.03cm}
  \caption{Comparisons of different algorithm parameters for pFedDef implementation on CIFAR-10 regarding impact on performance and robustness. High robustness against grey-box attacks can be achieved with low overhead parameter settings.}
  \label{fig:group_plot_3}
\end{figure*}

\subsection{Extended Numerical Analysis Setup} \label{sec:data-split}

All experiments in the main body and appendix are carried out on an AWS EC2 instance of type g4dn.xlarge. These instance types have NVIDIA GPUs using NVIDIA libraries such as CUDA.

\textbf{Non-i.i.d. Data Distribution.} The data distribution process across clients is taken from \citet*{fedMTL21}. When dividing data across clients during experiments, the parameter $\beta>0$ impacts how the data is distributed. The data division process begins with an assumption $M$ underlying distributions, identical to the set up of FedEM. The underlying distributions are constructed by having each label in the data set divided in a i.i.d. manner into one of the distributions. Afterwards, data points are mapped from each distribution to all clients using the Dirichlet distribution, which takes $\beta$ as an input parameter. When $\beta$ is a low value, data is more non-i.i.d. across clients as there is higher variance between clients for the number of data points assigned from a specific underlying distribution. When $\beta$ is a higher value, clients tend to have a similar number of data points from each underlying distribution compared to other clients, making the global data distribution more i.i.d.. For all experiments in the paper, the number of underlying distributions assumed is $M=3$. The impact of different settings of data distribution on federated learning and pFedDef are analyzed in Section \ref{sec:exp}.

\subsection{Number of Clients Participating in Federated Learning}

Federated learning (including personalized learning) is inherently scalable as a distributed learning scheme.
The performance of pFedDef and federated adversarial training (FAT) is analyzed for different numbers of learning clients in Figure \ref{fig:client_count}.  Both pFedDef and FAT observe small increases in test accuracy as the number of clients increase, due to a more diverse set of data becoming available for the learning task.
pFedDef especially scales well showing high robustness (Adv. Acc.) even with a small number of clients. The robustness propagation of algorithm \ref{alg:robust_prop} scales linearly with the number of clients.

\subsection{Overhead of pFedDef Training Parameters}\label{sec:overhead}

To observe the impact of pFedDef parameters on robustness and overhead, different values of parameters $G$ (desired adversarial data proportion), $Q$ (adversarial data set update frequency) and $K$ (number of PGD steps) are analyzed. Increasing each of these parameters increases overhead with respect to the adversarial training data generation. We only show results on the CIFAR-10 data set to conserve space. 
We note that targeted attacks presented in this analysis are created with the intent of altering the classification of a data point to a specific label. Untargeted internal evasion attacks are created following the method introduced in Section \ref{sec:transferability}.

Increasing both adversarial data proportion $G$ in Figure \ref{fig:G} and number of PGD steps $K$ in Figure \ref{fig:K} has similar effects on pFedDef's test accuracy and robustness. As both values are increased, test accuracy gradually decreases, while robustness against untargeted and targeted attacks increases. However, both the changes in test accuracy and robustness become minimal as $G$ exceeds 0.3 and $K$ exceeds 5. Thus, we can achieve high robustness with lower values of $G$ and $K$ that reduce algorithm overhead. Furthermore, robustness can be achieved when system resources $R_c$ are constrained for many clients due to resource propagation, as seen in Figure \ref{fig:cifar_prop}.

In Figure \ref{fig:Q}, as the number of rounds between adversarial data set updates ($Q$) increases, the test accuracy increases, indicating that the negative effect of adversarial training on test accuracy is less prevalent. In contrast, robustness against targeted and untargeted attacks initially increases and subsequently decreases as $Q$ is increased. Robustness is reduced when the value of $Q$ is too small as the training set is altered too quickly compared to the neural network parameters that are trained. Overall, the pFedDef algorithm achieves relatively high robustness given lower values of $G$ and $K$ and infrequent updates $Q$ to the adversarial data set, reducing the overhead of the adversarial training procedure.

\end{document}